\documentclass[sigconf,nonacm]{acmart}

\usepackage{graphicx,graphics}
\usepackage[utf8]{inputenc} 
\usepackage[T1]{fontenc}    
\usepackage{hyperref}       
\usepackage{url}            
\usepackage{booktabs}       
\usepackage{nicefrac}
\usepackage{wrapfig}
\usepackage{microtype}      
\usepackage{wrapfig,lipsum}
\usepackage{amsthm,mathrsfs,amsmath}
\usepackage{paralist}
\usepackage{tcolorbox}
\usepackage{pifont}
\usepackage{xargs}
\usepackage{dsfont}
\usepackage{caption}
\usepackage{subcaption} 
\usepackage{upgreek}
\usepackage{mathrsfs}
\usepackage{enumitem}
\usepackage{aliascnt}
\usepackage{multirow}
\usepackage[noend]{algpseudocode}  
\usepackage{algorithm}
\usepackage{algorithmicx}
\usepackage{bbm}
\usepackage{bm}
\usepackage{svg}
\usepackage{makecell}
\usepackage{xspace}
\graphicspath{{figures/}}
\usepackage[capitalize,noabbrev]{cleveref}
\theoremstyle{plain}
\newtheorem{theorem}{Theorem}[section]
\newtheorem{proposition}[theorem]{Proposition}

\theoremstyle{definition}
\newtheorem{definition}[theorem]{Definition}

\crefformat{assumption}{{\textbf{H}}#2#1#3}
\theoremstyle{remark}
\newtheorem{remark}[theorem]{Remark}
\newtheorem{example}[theorem]{Example}

\definecolor{aurometalsaurus}{rgb}{0.43, 0.5, 0.5}
\definecolor{britishracinggreen}{rgb}{0.0, 0.26, 0.15}
\definecolor{burntumber}{rgb}{0.54, 0.2, 0.14}
\definecolor{cobalt}{rgb}{0.0, 0.28, 0.67}
\definecolor{bulgarianrose}{rgb}{0.28, 0.02, 0.03}
\definecolor{ceruleanblue}{rgb}{0.16, 0.32, 0.75}

\newcommand{\B}{\boldsymbol}
\newcommand{\ourmethod}{\texttt{SPARTA}\xspace}
\newcommand{\averagedtrue}{\texttt{Oracle Mask--(non-private)}\xspace}
\newcommand{\averagednoisy}{\texttt{DP-SGD Gradients}\xspace}

\newcommand{\rowgrouping}{\texttt{row-grouping}\xspace}
\newcommand{\mpweights}{\texttt{MP}~\cite{cvpr}\xspace}
\newcommand{\lr}{$\texttt{lr}$\xspace}
\newcommand{\classifierlr}{$\texttt{classifier\_lr}$\xspace}

\newcommand{\cA}{\mathcal{A}}

\newcommand{\cG}{\mathcal{G}}
\newcommand{\cN}{\mathcal{N}}

\newcommand{\cD}{\mathcal{D}}

\newcommand{\p}{{\rm I}\kern-0.18em{\rm P}}

\newcommand{\group}[1]{\bgroup\small\sffamily\noindent\color{green}{#1}\egroup}  
\newcommand{\new}[1]{\bgroup\small\sffamily\noindent\color{cobalt}{#1}\egroup}
\newcommand{\old}[1]{\bgroup\small\color{gray}{#1}\egroup}  

\definecolor{darkgreen}{RGB}{0,128,0}
\renewcommand{\algorithmiccomment}[1]{{\small\textbf{\textcolor{darkgreen}{// \texttt{#1}}}}}
\newcommand{\pluseq}{\mathrel{+}=}

\hypersetup{colorlinks = true, urlcolor = blue, bookmarksopen = true}

\newcommand{\R}{\mathbb{R}}

\newcommand{\topk}{\operatorname{Top}-k}
\newcommand{\argtopk}{\operatornamewithlimits{\arg\topk}}

\newcommand{\1}{\mathbb{I}}

\newcommand{\abs}[1]{\left\lvert{#1}\right\rvert}

\AtBeginDocument{%
  }

\setcopyright{acmlicensed}
\copyrightyear{2025}
\acmYear{2025}
\acmConference[KDD'25]{}{August 03--07,2025}{Toronto, ON}

\author{Mehdi Makni}
\orcid{0009-0009-7573-8775}
\affiliation{%
  \institution{Massachusetts Institute of Technology}
  \city{Cambridge}
  \state{MA}
  \country{USA}
}
\email{mmakni@mit.edu}

\author{Kayhan Behdin}
\orcid{}
\affiliation{%
  \institution{Massachusetts Institute of Technology}
  \city{Cambridge}
  \state{MA}
  \country{USA}
}
\email{behdink@mit.edu}

\author{Gabriel Afriat}
\orcid{0000-0002-9730-5403}
\affiliation{%
  \institution{Massachusetts Institute of Technology}
  \city{Cambridge}
  \state{MA}
  \country{USA}
}
\email{afriatg@mit.edu}

\author{Zheng Xu}
\orcid{}
\affiliation{%
  \institution{Google Research}
  \city{Mountain View}
  \state{CA}
  \country{USA}
}
\email{xuzheng@google.com}

\author{Sergei Vassilvitskii}
\orcid{}
\affiliation{%
  \institution{Google Research}
  \city{New York}
  \state{NY}
  \country{USA}
}
\email{sergeiv@google.com}

\author{Natalia Ponomareva}
\orcid{}
\affiliation{%
  \institution{Google Research}
  \city{New York}
  \state{NY}
  \country{USA}
}
\email{nponomareva@google.com}

\author{Hussein Hazimeh}
\orcid{0000-0003-4501-0678}
\affiliation{%
  \institution{Google Research}
  \city{New York}
  \state{NY}
  \country{USA}
}
\email{hazimeh@google.com}

\author{Rahul Mazumder}
\orcid{0000-0003-1384-9743}
\affiliation{%
  \institution{Massachusetts Institute of Technology}
  \city{Cambridge}
  \state{MA}
  \country{USA}
}
\email{rahulmaz@mit.edu}

\begin{document}

\title{\ourmethod: An Optimization Framework for Differentially Private Sparse Fine-Tuning}

\begin{abstract}
Differentially private stochastic gradient descent (DP-SGD) is broadly considered to be the gold standard for training and fine-tuning neural networks under differential privacy (DP). 
With the increasing availability of high-quality pre-trained model checkpoints (e.g., vision and language models), fine-tuning has become a popular strategy. However, despite recent progress in understanding and applying DP-SGD for private transfer learning tasks, significant challenges remain -- most notably, the performance gap between models fine-tuned with DP-SGD and their non-private counterparts. 
Sparse fine-tuning on private data has emerged as an alternative to full-model fine-tuning --- recent work has shown that privately fine-tuning only a {\em small subset} of model weights and keeping the rest of the weights fixed can lead to better performance. In this work, we propose a new approach for sparse fine-tuning of neural networks under DP.
Existing work on private sparse finetuning often used fixed choice of trainable weights (e.g., updating only the last layer), or relied on public model's weights to choose the subset of weights to modify. Such choice of weights remains suboptimal. In contrast, we explore an optimization-based approach, where our selection method makes use of the private gradient information, while using off the shelf privacy accounting techniques.  
Our numerical experiments on several computer vision models and datasets show that our parameter selection method leads to better prediction accuracy, compared to full-model private fine-tuning or existing private sparse fine-tuning approaches.
\end{abstract}



\begin{teaserfigure}
  \centering
  \includegraphics[width=0.95\textwidth]{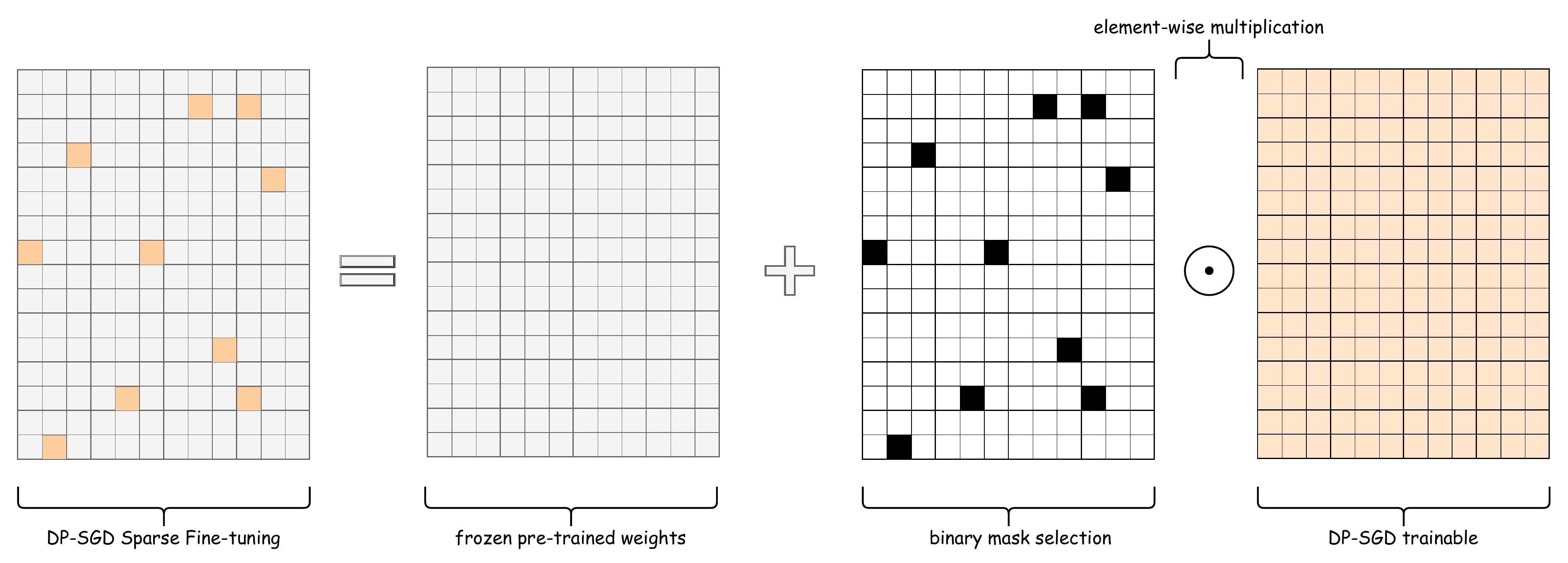}
  \caption{\ourmethod finds a good mask in this formulation and shows Sparse Fine-Tuning improves DP-SGD dynamics.}
  \label{fig:teaser}
\end{teaserfigure}

\maketitle

\section{Introduction}
\label{sec:introduction}

Real-world datasets often contain sensitive or personal information, such as financial or healthcare information, that needs to be protected from bad actors. In fact, the responsibility of protecting such sensitive information has been encoded into the law by regulations such as the EU’s General
Data Privacy Regulation (GDPR) and the California Consumer Privacy Act. This poses challenges for modern machine learning systems that depend on large-scale datasets to deliver good predictive performance. 
As a result, recent years have seen a growing interest in privacy-preserving machine learning.
Among numerous privacy-preserving machine learning frameworks~\citep{privacy2,privacy1}, differential privacy (DP,~\citep{dwork2006differential}) is widely accepted as the gold standard for providing rigorous privacy guarantees~\citep{dwork2014algorithmic}. Loosely speaking, DP ensures that an adversary cannot learn about any individual training data sample by examining the learned model.  A significant body of work has been dedicated to studying the theoretical and algorithmic aspects of machine learning under DP~\citep{dwork2014algorithmic, dpfy}. 

Modern computer vision systems make use of the immense expressive power of deep neural networks (DNNs). The main workhorse for training DNNs under DP is differentially private stochastic gradient descent (DP-SGD,~\citep{abadi2016deep}). DP-SGD introduces two major modifications to an SGD algorithm: per-sample gradient clipping and noise injection. Gradient clipping is implemented in DP-SGD to limit the influence of each training sample on the resulting model, and noise injection ensures that information about any individual is sufficiently obscured in the final model. Here, we assume that each user contributes a single example to the training data, which ensure example-level DP is equivalent to user-level DP. Unfortunately, these two steps (clipping and noising) can result in significant predictive performance degradation compared to non-private SGD~\citep{kurakin2022toward,dpmoredata,papernot2019making}, especially for models with a large number of parameters---DP-SGD suffers from the curse of dimensionality~\citep{normd1,dpsize,yu2020not}. For example, everything else being equal, the expected norm of the noise injected in DP-SGD is proportional to $\sqrt{d}$, where $d$ is the number of model weights. Computer vision tasks often rely on DNNs with the number of parameters in tens or hundreds of millions~\citep{resnet,wrn,deit}, where the application of DP-SGD might come with significant loss of model utility.

In practice, large models that are trained on publicly available datasets are often fine-tuned on private data to perform well on specific tasks. Private fine-tuning of pre-trained networks has been shown to perform well in practice~\citep{ganesh2023public,li2022does}. Since publicly pre-trained models commonly perform well on general machine learning tasks, it is natural to ask if every parameter of the pre-trained model has to be fine-tuned to obtain a good performance on the private data task. Fine-tuning a small subnetwork of a large model, or sparse fine-tuning, can help to reduce the utility loss of DP-SGD, by reducing the amount of the noise that needs to be injected in DP-SGD updates.~\citet{cvpr} were one of the first to demonstrate private and sparse fine-tuning of large pre-trained models can perform comparably to full-model training and/or fine-tuning. A long line of follow-up work~\citep{firstlastwrn,deepmindft,mehta2022large,preprune} has studied the problem of sparse fine-tuning under differential privacy, and developed several algorithmic approaches that can perform better than whole network fine-tuning. 

Parameter Efficient Fine-Tuning (PEFT) is a way to reduce the number of trainable parameters while preserving the model's utility. In general this line of work has two branches: The first one is focused on freezing the existing model weights and introducing additional parameters to be tuned. This increases the model size and can make the inference slower, but has been shown to perform extremely well. Examples of such work are LoRA~\cite{hu2022lora} tuning, Prompt tuning approaches~\cite{jia2022visual}, Adapter tuning~\cite{he2021effectiveness}, etc. This line of work is very popular in the field of Large Language models. The second group of PEFT methods concentrates instead on choosing the weights to update among the already pre-trained and existing weights. Examples of this work include tuning only the last layer, bias-term only fine-tuning, and others. While there has been work showing that the first group of PEFT for DP-SGD improves the utility (e.g. \citet{dplora1, dplora2} show that LoRA does better than full fine-tuning), in this work we explore if the same holds for the second group. Our results suggest that choosing the weights to fine-tune with DP SGD is beneficial and confirms the folklore knowledge that fewer but "better" parameters for DP-SGD results in better utility. 

It is also worth noting that we consider DP-SGD as a baseline for our experiments (showcasing improved utility by introducing sparse fine-tuning), but our proposed method can be applied to newer generation algorithms for DP like DP-MF~\cite{choquette2024amplified}. 

In what follows, we discuss our approach and contributions, and then review the related work.

\noindent\textbf{Summary of approach and contributions:}
We propose \ourmethod: \underline{S}parse \& \underline{P}riv\underline{A}te \underline{R}ow--gradient \underline{T}hresholding \underline{A}lgorithm,
an end-to-end sparse differentially private fine-tuning optimization framework. \ourmethod is modular and can be adapted to existing DP training libraries---we present an implementation of \ourmethod in Opacus~\citep{opacus}. Our experiments on a wide range of fine-tuning tasks show how \ourmethod improves the utility in DP fine-tuning over existing fine-tuning approaches, either sparse or dense. We build our framework on three major pillars:

\noindent\emph{Optimization formulations:} Prior work has shown that the choice of the fine-tuning subnetwork plays an important role. Therefore, starting from the first principles, we present an optimization formulation, to simultaneously  (1) choose the fine-tuning subnetwork; and (2) fine-tune the weights of the selected subnetwork. Our optimization-based approach to selecting the fine-tuning subnetwork is different from the previous work that  used a fixed selection of weights to tune (such as last/first layer) or relied on publicly available model weights.

\noindent\emph{Approximate algorithms:} Our joint selection and fine-tuning optimization problem involves a combinatorial structure which can be computationally challenging. To this end, we propose a first-order approximate algorithm to obtain good feasible solutions to our optimization. Our numerical experiments show that our algorithm is able to identify subnetworks such that private fine-tuning on them outperforms private fine-tuning of the whole network or commonly studied fixed subnetworks (such as last layer, or first-and-last layer). 

\noindent\emph{Privacy accounting:} Our approximate algorithm makes use of the (private) fine-tuning data. We show that existing off-the-shelf privacy accountants can be used to keep track of the privacy budget of our joint selection and training procedure. In particular, we show that the privacy cost of our subnetwork selection is the same as the cost of one additional epoch of DP-SGD, a numerically insignificant overhead in practice.

\textit{Our contributions} can be summarized as follows: (1) We present an optimization-based approach to jointly select and fine-tune subnetworks of large pre-trained models under differential privacy. As our optimization formulation is computationally challenging to solve, we present an approximate algorithm to obtain good solutions to our joint optimization framework. (2) Given that our subnetwork selection uses private gradient information, we study the privacy overhead of our selection method. We show the privacy overhead of our proposal is the same as one epoch of DP-SGD. (3) Our numerical experiments on several computer vision fine-tuning tasks show that our selection method outperforms existing sparse DP fine-tuning methods  (e.g., last or first-and-last layer fine-tuning) and whole network fine-tuning.

\paragraph{Related Work: Sparse DP fine-tuning}
In this paper, we focus on sparse DP fine-tuning which has received significant attention in the recent years. Most existing approaches rely on the pre-trained network for the choice of trainable subnetworks, or use fixed subnetworks. For example,~\cite{cvpr} choose to train the classifier and normalization layers, and a fixed percentage of the weights that have the largest magnitude in the pre-trained model. \cite{mehta2022large,deepmindft} discuss fine-tuning only the last layer of the network, while~\cite{firstlastwrn} proposes to fine-tune the first and last layer of the pre-trained network. \cite{preprune} proposes to first reduce the size of the pre-trained network through pruning the less important weights. They also propose to only update a subnetwork of the pruned network for further dimension reduction. Another approach is by~\cite{bu2022differentially}, which propose to only fine-tune the bias terms. Overall, approaches discussed here fall under two broad classes: (1) They use a fixed selection of subnetworks, such as last layer; or, (2) To select the trainable subnetwork, they make use of publicly available model weights, either from the pre-trained network or the weights that have been updated via DP-SGD and can be released publicly. As neither of these approaches use private information, they broadly do not need additional privacy accounting. However, not making use of the private information can potentially lead to subnetwork selections that result in sub-optimal prediction performance---as demonstrated by~\cite{firstlastwrn}, the performance of sparse DP fine-tuning can improve with a better choice of the trainable subnetwork.

\section{Preliminaries}
Before discussing our method, let us briefly review some notation and important concepts that we will use throughout the paper.

\noindent\textbf{Notation:} Let $\B W\in\R^d$ parameterize the network weights, where $d$ is the network dimension. We assume that the private fine-tuning dataset contains $n$ observations, and  let $\mathcal{L}_i(\B W)$ for $i\in[n]$ denote the fine-tuning loss, corresponding to the $i$-th data point. Define $\mathcal{L}(\B W)= \sum_{i=1}^n \mathcal{L}_i(\B W)/n$ to be the overall fine-tuning loss. For $\B{v}\in\R^d$ and $k\geq 1$, we let $\argtopk(\B v)\in\{0,1\}^d$ be such that the $j$-th coordinate of $\argtopk(\B v)$ is equal to one, if and only if $v_j$ is among the coordinates of $\B v$ with $k$ largest magnitudes. We also let $\1(\cdot)$ denote the indicator function; for example, $\1(x=y)=1$ if $x=y$ and $\1(x=y)=0$ otherwise. Also, for $\B{g}\in\R^d$, denote $|\B{g}|\in\R^d$ the vector of elementwise absolute value entries of $\B{g}$; that is $|\B{g}| = (|g_1|, \dots, |g_d|)$. Finally, let $\B I$ denote the identity matrix (of size $d$), and $\cN(0,\sigma^2 \B{I})$ a sample from a $d$-dimensional Gaussian distribution with zero mean and covariance $\sigma^2 \B I$.\\

\noindent\textbf{Differential Privacy:}
We present a formal definition of DP, which relies on the notion of {\em neighboring} datasets that typically differ in one record (See the discussion in ~\cite{dpfy} for more details). 

\begin{definition}[DP, \citet{dwork2006differential}, \cite{abadi2016deep}]\label{dp-def}
    Given the privacy parameters $(\varepsilon, \delta) \in \R^+ \times \R^+$, a randomized algorithm $\cA(\cdot)$ is said to satisfy the $(\varepsilon, \delta)$-DP property if
    $$\p(\cA(\cD)\in K)\leq e^{\varepsilon}\p(\cA(\cD')\in K)+\delta.$$
for any measurable event $K \in \text{range}(\cA)$ and for any pair of neighboring datasets $\cD$ and $\cD'$.
\end{definition}
We note that in Definition~\ref{dp-def}, the probability is taken over the randomness of the algorithm $\cA$. Stronger DP guarantees are measured by smaller $(\varepsilon, \delta)$ values. A large body of work has been dedicated to developing algorithms for learning under DP~\citep{dwork2014algorithmic}. In this work, we focus on DP-SGD (although our proposed approach can be applied on top of other DP-learning algorithms like DP-MF~\cite{choquette2024amplified}.) \\

\noindent\textbf{DP-SGD:} 
\citet{abadi2016deep} propose DP-SGD as a modification of stochastic gradient descent for DP learning. In DP-SGD, stochastic data minibatches are used to update the model weights. For a fixed sampling probability $0<q\leq 1$, at iteration $t$, a random minibatch (or lot) $B_t\subseteq[n]$ is drawn, such that each data point is chosen to be in $B_t$ with probability $q$ independent from other data points. Let $\B W^t$ denote the network weights at iteration $t$ of DP-SGD, and let
$\B g^{t,i}=\nabla \mathcal{L}_i(\B W^t)$
be the gradient of the loss of the $i$-th data point, at iteration $t$. For every data point $i\in B_t$, the per-sample clipped gradient is calculated as 
$$\tilde{\B g}^{t,i} = \B g^{t,i}/\max\left\{1,\frac{\|\B g^{t,i}\|_2}{C}\right\}.$$
where $C>0$ is the clipping threshold. This step ensures the influence of $i$-th sample is limited in the resulting model. The noisy batch gradient is then obtained as
\begin{equation}\label{noisygrad}\tilde{\B g}^t = \frac{1}{qn}\left(\sum_{i\in B_t}  \tilde{\B g}^{t,i} + \cN(0,\sigma^2 C^2 \B{I})\right).\end{equation}
for some suitable noise multiplier $\sigma>0$. The Gaussian noise is injected here to ensure the overall mechanism remains private. Finally, the weights are updated as 
$\B W^{t+1} = \B W^t - \eta_t \tilde{\B g}^t$
for some learning rate $\eta_t>0$. The work of \citep{abadi2016deep} discuss how to calculate the overall privacy guarantees given the values of $q,\sigma$.\\

\noindent\textbf{Subsampled Gaussian Mechanism (SGM):} The noisy gradient update~\eqref{noisygrad} of DP-SGD is a special case of an SGM, defined below.

\begin{definition}[SGM,~\citep{rdp1}]\label{sgm}
 Let $f$ be a function mapping subsets of $[n]$ to $\R^d$. Suppose $f$ has a global sensitivity of $C>0$: $\|f(S)-f(S')\|_2\leq C$ for any two adjacent $S,S'\subseteq[n]$. The Sampled Gaussian Mechanism (SGM), parameterized with the sampling rate $0 < q \leq 1$ and the
noise variance $\sigma^2>0$ is defined as
$$SG_{q,\sigma,C}(S)= f(S) + \cN (0, C^2\sigma^2\B I),$$
where each element of $S$ is sampled independently at random with probability $q$ without replacement.
\end{definition}
Given this definition, DP-SGD can be thought of as a composition of a series of SGMs.
Although \cite{abadi2016deep} discusses privacy accounting for DP-SGD, as it turns out, similar privacy accountants can be designed for general SGMs. For example, the Opacus'~\citep{opacus} default privacy accountant~\citep{prv-opacus} can keep track of the privacy budget of composition of SGMs (of which, DP-SGD is a special case).


\section{An Optimization Formulation for Sparse Fine-Tuning}
\label{sec:formulation}

In this section, we discuss a general optimization framework for sparse fine-tuning of DNNs. We first study the non-private setting to develop our framework, then we discuss how our framework can be applied under DP constraints.
\subsection{Non-Private Formulations}
We start with the non-private setting. Suppose $\B W_{\text{old}}$ denotes the current model weights. This can be the pre-trained network, or a slightly fine-tuned model (e.g., for a few epochs). We let $\B{m}\in\{0,1\}^d$ be a mask that indicates which model weights will get updated. If $m_i=1$ for $i\in[d]$, the $i$-th weight receives gradient updates and gets fine-tuned. If $m_i=0$, the $i$-th weight stays the same as the current weight. Thus, we can parameterize the fine-tuned DNN as $\B{W}_{\text{old}}+\B{m}\odot \B{W}$ where $\odot$ denotes the Hadamard product, and $\B{W}$ are the (fine-tuned) weights we optimize over. We would like to minimize the fine-tuning loss, $\mathcal{L}(\B{W}_{\text{old}}+\B{m}\odot \B{W})$. This minimization however, is jointly over $\B{m}$ (i.e., which weights to fine-tune) and $\B{W}$ (i.e., what the fine-tuned weights should be).

For additional modeling flexibility, let us assume we select groups of model parameters to be trainable in our subnetwork selection procedure. As we discuss later, this group selection structure helps when we study the private setting. Formally, suppose the network weights are partitioned into $q\geq 1$ non-intersecting groups, given as $\cG_1,\cdots,\cG_q\subseteq [d]$. For every group $\cG_i$, we either set all the weights belonging to this group to be trainable, or none of them to be trainable. We assume that this grouping is fixed and is given a priori. In the extreme case, each group can contain a single model weight, implying that we can set individual weights to be trainable or not. We let $\B z\in\{0,1\}^q$ be the indicator that specifies if a group of weights is trainable or not. If $z_j=1$ for some $j\in[q]$, then all the weights in $\cG_j$ are set to be trainable, $m_i=1$ for all $i\in\cG_j$. We can also control how many groups of variables are set to be trainable by constraining $\sum_{j=1}^q z_j\leq k$ where $k$ is the budget on the number of trainable groups. Thus, we can define the set of all possible masks with the budget of $k$ trainable groups as
\begin{align}\label{deltadef}
    \Delta = \bigg\{&(\B m,\B z):\B m\in\{0,1\}^d,\B z\in\{0,1\}^q,\\
    &\sum_{j=1}^q z_j \leq k,~~m_i\leq z_j~~\forall i\in\cG_j, j=1,\cdots,q\bigg\} \notag.
\end{align}
Using these principles, we arrive at our optimization-based approach for subnetwork selection:

\begin{equation}\label{general-1}
    \min_{\B W, \B{m}, \B z} ~~ \mathcal{L}(\B W_\text{old} + \B{m} \odot \B{W})~~~~
     \text{s.t.}  ~~~~ (\B m,\B z)\in\Delta.
\end{equation}

Here, we are jointly selecting the weights and the (sparse) mask to minimize the fine-tuning loss. The constraint  $(\B m,\B z)\in\Delta$ ensures the masks we obtain from~\eqref{general-1} are sufficiently sparse.

\begin{remark}[Differences from neural network pruning] We note that Problem~\eqref{general-1} is different from the DNN pruning problem. In DNN pruning~\citep{obd,obs,obc,chita}, the goal is to choose a sparse subnetwork that leads to small training error, when every other network weight is set to zero. In Problem~\eqref{general-1} we do not set any weights to zero, rather, we \emph{freeze} some weights and do not update them during training.
\end{remark}

Problem~\eqref{general-1} is non-convex and has a combinatorial structure. This makes solving it computationally challenging. Therefore, we use a linear approximation to $\mathcal{L}(\B W_{\text{old}}+\B m\odot \B{W})$, and we add an additional regularization term, as this approximation is only valid locally, to ensure that the fine-tuned weights $\B W_{\text{old}}+\B m\odot \B{W}$ do not move too far away from the current weights $\B W_{\text{old}}$. Specifically, letting $\tilde{\B W}=\B{W}_{\text{old}}+\B m\odot \B W$, we approximate
using a second-order Taylor expansion of the loss function $\mathcal{L}$. To this end, for any $\tilde{\B W}$ we can write
\begin{multline}\label{localquadratic}
    \mathcal{L}(\tilde{\B W}) \approx  \mathcal{L}({\B W}_{\text{old}}) + (\tilde{\B W}-{\B W}_{\text{old}})^T\nabla \mathcal{L}(\B W_{\text{old}})
    + \frac{\tau}{2} \|\tilde{\B W}-{\B W}_{\text{old}}\|_2^2.
\end{multline}
where $\nabla \mathcal{L}$ denotes the gradient of the loss function, and $\tau>0$  is the regularization coefficient we discuss later.

Under this approximation, by substituting~\eqref{localquadratic} into~\eqref{general-1}, we obtain an approximation to Problem~\eqref{general-1} as
\begin{multline}\label{diagonal-problem}
  \min_{ \B W, \B{m}, \B z}  ~~~~\B g^T (\B{m}\odot \B{W})  + \frac{\tau}{2} \|\B{m}\odot \B{W}\|_2^2 ~~~~ \text{s.t.}  ~~~~ (\B m,\B z)\in\Delta.
\end{multline}
where $\B g=\nabla \mathcal{L}(\B W_{\text{old}})$. 

We now discuss how good solutions to Problem~\eqref{diagonal-problem} can be found. Note that Problem~\eqref{diagonal-problem} has the same minimizer as Problem~\eqref{diagonal-upper} given as:
\begin{align}\label{diagonal-upper}
   \min_{\B W, \B{m}, \B z} & ~~ \left\Vert \B{m}\odot \B{W} + \B{g}/\tau \right\Vert_2^2 ~~~~
     \text{s.t.}  ~~~~ (\B m,\B z)\in\Delta.
\end{align}

Upon inspection, we note that the optimal solution to Problem~\eqref{diagonal-upper} can be found in closed-form.
Define $\B v\in\R^q$ such that
\begin{equation}\label{vdef}
    v_j = \sum_{i\in\cG_j} g_i^2~~\forall j\in [q].
\end{equation}
Then, it is easy to see that
   \begin{align}
       \hat{\B W} &= -\frac{\B{g}}{\tau},~~~~ \hat{\B z} = \argtopk(\B v), \notag \\\hat{m}_i&=\sum_{j\in[q]}\1(\hat{z}_j=1,i\in\cG_j)~~\forall i\in[d].
   \end{align}

However, under privacy constraints, it is more appealing to consider an upper bound to Problem~\eqref{diagonal-upper} in Problem~\eqref{diagonal-upper-norm1}---we will discuss privacy cost accounting for Problem~\eqref{diagonal-upper-norm1} in Section~\ref{sec:ourmethod}:
\begin{align}\label{diagonal-upper-norm1}
   \min_{\B W, \B{m}, \B z} & ~~ \left\Vert \B{m}\odot \B{W} + \B{g}/\tau \right\Vert_1^2  ~~~~
     \text{s.t.}  ~~~~ (\B m,\B z)\in\Delta.
\end{align}
Note that the optimal solution to Problem~\eqref{diagonal-upper-norm1} changes the \textit{group scoring function} $v_j$ to be the $\ell_1$-norm of the gradient of the group $j \in [q]$. The optimal solutions to Problem~\eqref{diagonal-upper-norm1} are therefore given by
\begin{align}\label{easy-update}
    v_j &= \sum_{i\in\cG_j}|g_i|~~\forall j\in [q].\notag\\
    \hat{\B W} &= -\frac{\B{g}}{\tau},~~~~ \hat{\B z} = \argtopk(\B v), \notag \\\hat{m}_i&=\sum_{j\in[q]}\1(\hat{z}_j=1,i\in\cG_j)~~\forall i\in[d].
\end{align}
In other words, to solve Problem~\eqref{diagonal-upper-norm1}, we need to sort the coordinates of $\B v$ based on their magnitude, and select to train the groups corresponding to the $k$ largest ones. We can use the update in~\eqref{easy-update} iteratively to carry out sparse fine-tuning, by updating $\B W_{\text{old}}\gets \B W_{\text{old}} + \hat{\B m}\odot \hat{\B W}$. Here, $1/\tau$ can be thought of as the step size. In fact, if we allow all groups of variables to be trainable, that is, $k=q$, and set $\eta_t=1/\tau$, then~\eqref{easy-update} simplifies to ordinary (S)GD.

\subsection{Privatizing~\eqref{easy-update}: A first attempt}
The update~\eqref{easy-update}, however, uses the private gradients $\B{g}$ and therefore does not satisfy differential privacy.  A first natural idea would be to use the noisy and clipped gradients available from DP-SGD in~\eqref{easy-update} to ensure the whole mechanism remains private. However, especially in high privacy regimes (small $\varepsilon$), the norm of noise can dominate the norm of the gradient --- the updates~\eqref{easy-update} would not perform better than randomly choosing which model weights to update. We demonstrate this below by a numerical example.

\begin{example}\label{example1}
Similar to the ablation study conducted in \cref{sec:ablation}, we consider DP fine-tuning on the CIFAR10 dataset ~\citep{cifar} of the ResNet18 network~\citep{resnet} that has been pre-trained on CIFAR100~\citep{cifar} with $\varepsilon=1,\delta=10^{-5}$. We use 5 independent runs for the error bars. We use~\eqref{easy-update} to obtain a mask for fine-tuning where $\B g$ is substituted with DP-SGD gradients averaged over an epoch. That is, $\B g\approx \sum_{t=1}^{T_b}\tilde{\B g}^t/T_b$ where $T_b$ is the number of iterations in an epoch and $\tilde{\B g}^t$ is in~\eqref{noisygrad}. We do not use grouping, $\cG_i=\{i\}$ for $i\in[d]$ and we allow $20\%$ of the weights to be trainable by choosing $k=0.2d$. The additional details of this procedure can be found in Algorithm~\ref{algo:ex1} in the Appendix.  We carry out sparse and private fine-tuning with DP-SGD on the resulting mask.  At the end, we achieve an out-of-sample accuracy of $77.89\pm(0.05)\%$. 
However, under the same setting, if we fine-tune the whole network with DP-SGD using a completely random mask, we achieve an out-of-sample accuracy of $77.84\pm(0.17)\%$.
\end{example}
As we observe in Example~\ref{example1}, using DP-SGD gradients for mask selection does not result in an improvement over a random choice of the mask. Motivated by this, we introduce \ourmethod, that performs the privacy accounting in a more clever fashion and selects a mask that results in better DP-SGD dynamics --- under the same setup as Example~\ref{example1}, \ourmethod achieves an out-of-sample accuracy of $78.81\pm(0.12)\%$.

\section{\ourmethod: Sparse and Private Fine-Tuning}\label{sec:ourmethod}
In this section, we introduce \ourmethod, our end-to-end sparse DP fine-tuning procedure. We discuss how the mask we obtain from our optimization-based approach can be released under privacy constraints.
The update in~\eqref{easy-update} shows that to select a particular group of model weights such as $\cG_j$ to be trainable, the corresponding coordinate of $\B v$ has to be large. This implies that the weights $i$ in group $\cG_j$ have to have large absolute gradients, \emph{on average}. In other words, the knowledge of $\B v$ suffices to obtain $\hat{\B m}$ in~\eqref{easy-update}. Luckily, as we discuss below, estimating the group aggregate information $\B v$ under privacy constraints is rather straight-forward.

To this end, suppose a data batch $B_t$ is selected at iteration $t$, and let 
$$\B{g}^{t,i}=\nabla\mathcal{L}_i(\B W^t).$$
be the gradient of a data point $i\in B_t$. Suppose we are using DP-SGD with the clipping constant of $C>0$ and the noise standard deviation $\sigma$. Define 
$$ \B{u}^{t,i} = |\B g^{t,i}|/\max\left\{1,\frac{\|\B g^{t,i}\|_2}{C}\right\}.$$
to be the absolute value of the clipped gradient. Let 

\begin{equation}\label{tildeut}
    \tilde{\B u}^t =\sum_{i\in B_t} \B{u}^{t,i} + \cN(0,\sigma^2C^2\B I).
\end{equation}
The vector $\tilde{\B u}_t$ is an estimate of the absolute value of a batch gradient that can be publicly released, achieved by clipping the absolute value of per-sample gradients and then adding Gaussian noise. In particular, if $C=\infty,\sigma=0$, we have $\tilde{\B u}^t=\sum_{i\in B_t} |\B{g}^{t,i}|$. Next, we consider the vector $\tilde{\B v}^t\in\R^q$ with

\begin{equation}\label{vtildet}
    \tilde{v}_j^t = \sum_{i\in\cG_j} \tilde{u}_i^t.
\end{equation}
which is a ``clipped and noisy'' version of the vector $\B v$ in~\eqref{vdef}, calculated on the batch $B_t$. 
 Consider Proposition~\ref{sgm-prop}.
\begin{proposition}\label{sgm-prop}
    The vector $\tilde{\B u}^t$ in~\eqref{tildeut} is an SGM as defined in Definition~\ref{sgm}.
\end{proposition}
Proposition~\ref{sgm-prop} shows that obtaining $\tilde{\B u}^t$ (and $\tilde{\B v}^t$ consequently) has the exact same privacy cost as a step of DP-SGD (with the clipping constant of $C$ and the noise parameter $\sigma$) if one uses a DP accountant that considers SGMs and their compositions, for instance,~\citep{prv-opacus} which is the default in Opacus. Therefore, we calculate and average $\tilde{\B v}^t$ over multiple batches $B_1,\cdots,B_{T_b}$ for some $T_b\geq 1$ to obtain an estimate of $\B{v}$ such as $\tilde{\B v}=\sum_{t=1}^{T_b} \tilde{\B v}^t/T_b$. For every batch $B_t$, we include the privacy cost of one DP-SGD step in the privacy accountant, which ensures each $\tilde{\B v}^t$ can be publicly released. As each $\tilde{\B v}^t$ and consequently 
$\tilde{\B v}$ can be publicly released,
 we replace $\B{v}$ with $\tilde{\B v}$ in~\eqref{easy-update} and obtain a mask:
 
   \begin{equation}\label{easy-update-dp}
       \hat{\B z} = \argtopk(\tilde{\B v}), ~~~~~ \hat{m}_i  =\sum_{j\in[q]}\1(\hat{z}_j=1,i\in\cG_j)~\forall i\in[d].
   \end{equation}
   The mask that is obtained from~\eqref{easy-update-dp} can be released publicly with no additional privacy cost, as it is a result of composition of SGMs.
   We note that, however, if we use some data batches to calculate $\hat{\B m}$, we cannot use the same data batches to calculate the DP-SGD gradients. Therefore, if we use $T_b$ batches of data to obtain a mask, we have to carry out $T_b$ fewer steps of DP-SGD to obtain the same DP guarantees as if we did not use any data to find a mask. 
   
      The procedure we use in practice is shown in Algorithm~\ref{algo:row-pruning}. Initially, we fine-tune the pre-trained model using DP-SGD for $T_0$ epochs, to make sure the model gradients are informative (note that the last-layer in vision tasks is randomly initialized so initial gradients may not be reliable for mask selection). Then, we stop DP-SGD for an epoch, and instead use the data batches to obtain a mask via~\eqref{easy-update-dp}. Next, we continue DP-SGD training on the subnetwork indicated by $\hat{\B m}$ for another $T-T_0-1$ epochs. As discussed above, this procedure has the same privacy guarantees as if we run DP-SGD for $T$ epochs. This is formalized in Proposition~\ref{prop-algo}.
      \begin{proposition}\label{prop-algo}
          Under privacy accountant~\citep{prv-opacus} (or any accountant studying SGMs), Algorithm~\ref{algo:row-pruning} has the same privacy guarantees as $T$ epochs of DP-SGD.
      \end{proposition}

\subsection{Reasoning behind algorithm design choices}
      
    We can now answer three important questions regarding our fine-tuning procedure:

\noindent\textbf{Why we use grouping:}
Our discussion above illustrates the benefits of using group selection. If there are multiple weights in $\cG_j$ for some $j\in[q]$, in order to calculate $\tilde{v}_j^t$ in~\eqref{vtildet}, we average multiple coordinates of $\tilde{\B u}^t$. Therefore, the noise in coordinates of $\tilde{\B u}^t$ cancel each other when calculating $\tilde{\B v}$---it is less costly to publicly release a measure that is aggregated over multiple model weights. We study the effect of grouping numerically in Section~\ref{sec:experiments}.

\noindent\textbf{Why we do not use DP-SGD gradients for mask selection:} We note that the noise cancellation effect would not happen if we use the absolute value of noisy and clipped gradients of DP-SGD in~\eqref{tildeut}, that is, defining $\tilde{\B u}^t$ as 
$$\tilde{\B u}^t = \bigg\vert \sum_{i\in B_t}  \tilde{\B g}^{t,i} + \cN(0,\sigma^2 C^2 \B{I}) \bigg\vert, $$
where $\tilde{\B g}^{t,i}$ is the clipped gradient.\\ 
\noindent\textbf{Why we use the approximation in~\eqref{diagonal-upper-norm1} rather than~\eqref{diagonal-upper}:} We note that under Problem~\eqref{diagonal-upper-norm1}, we need to release the absolute values of gradients (rather than the squared gradients that would be required for Problem~\eqref{diagonal-upper}), which is more straightforward to write as an SGM.
\subsection{Implementation Details}
\begin{figure}
    \centering
    \includegraphics[width=0.5\textwidth]{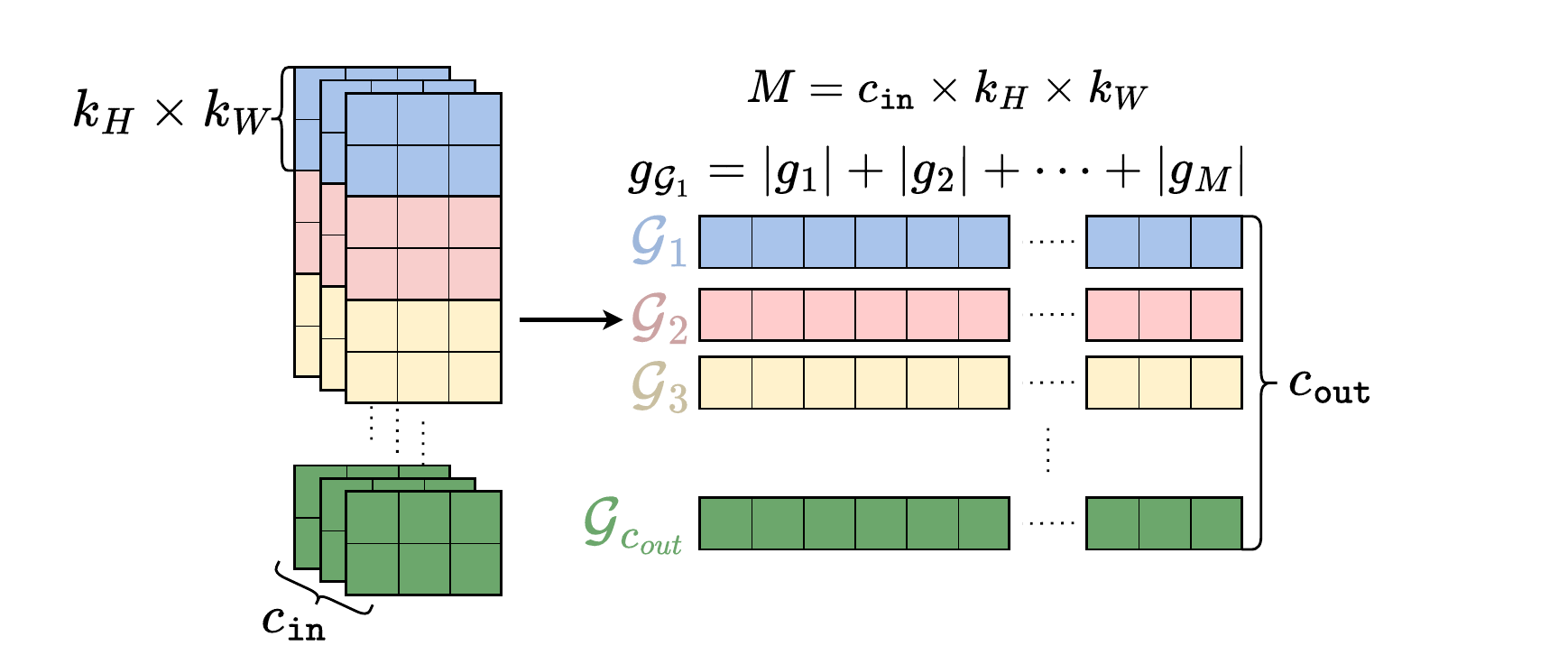}
    \caption{row grouping operation on a 2D convolutional Layer.}
    \label{fig:row_grouping_svg}
\end{figure}

Finally, we discuss some implementation details of \ourmethod.\\
\noindent\textbf{The choice of groups:}
The motivation of group selection discussed above does not impose any structure or constraints on how the groups are pre-chosen. In practice, we use \rowgrouping as the introduced row-structure can be exploited for efficiency in sparse fine-tuning (see ~\cref{main:sec-hardware}).
The \rowgrouping~is a per-layer grouping scheme that reshapes a high-dimensional tensor, representing layer weights, into a 2-dimensional tensor and selects rows of this matrix as groups.
This is illustrated in Figure~\ref{fig:row_grouping_svg} on a 2D Convolutional layer matrix with $c_\texttt{in}, c_{\texttt{out}}$ are the number of inputs and output channels respectively and $k_H, k_W$ the kernel dimensions.
\noindent\textbf{Per-layer sparsity budget:} We mostly discussed an overall sparsity budget for the number of trainable groups, given by $k$ in~\eqref{deltadef}. However, we note that our framework is flexible and can handle other types of sparsity constraints. In practice, we would like to ensure that a portion of each layer receives fine-tuning updates. Therefore, if we would like that at most $s\%$ of the mask to be nonzero, we select the mask so that for each layer at most $s\%$ of the weights are trainable. 

\begin{algorithm}[!ht]
    \small
    \setlength{\baselineskip}{1.2\baselineskip} 
    \renewcommand{\algorithmicindent}{1em} 
    \caption{\ourmethod}
    \label{algo:row-pruning}
    \begin{algorithmic}
        \State \textbf{Input:} $\B W_{\text{old}}$ (pre-trained weights), $T_0=10$ (mask finding epoch), $T=50$ (\#epochs)
           \State Train $\B W_{\text{old}}$ for $T_0$ epochs with DP-SGD. 
        \State \Comment{In practice, we do DP-BitFit during the first $T_0$ epochs.}
        \State $\tilde{\B u}=\B{0}$
        \For{each Poisson-sampled batch $B_t$ ($1$ epoch)}
                \State $\tilde{\B u} \pluseq \sum_{i\in B_t} |\B g^{t,i}|/\max\left\{1,\|\B g^{t,i}\|_2/C\right\} + \mathcal{N}(0, C^2\sigma^2 \B I) $ 
                \State\Comment{Private gradient absolute value update}
        \EndFor
        \State $\tilde{v}_j = \sum_{i\in\cG_j} \tilde{u}_i~~\forall j\in[q] $
        \State $\hat{\B z} = \argtopk(\tilde{\B v})$
        \State $\hat{m}_i=\sum_{j\in[q]}\1(\hat{z}_j=1,i\in\cG_j)~\forall i\in[d]$ 
        \State \Comment{Select and fix the sparse mask}
        \State DP-SGD training $\B{W}\mapsto \mathcal{L}(\B{W}_{\text{old}}+\hat{\B m}\odot \B{W} )$ for $T - T_0 - 1$ epochs.
        \State \Comment{Differentially Private Sparse fine-tuning.}
    \end{algorithmic}
\end{algorithm}

\section{Mask selection ablation Study}
\label{sec:ablation}

We perform additional ablation studies on the ResNet18 network~\citep{resnet} that has been pre-trained on CIFAR100~\citep{cifar}. We consider DP fine-tuning this network on CIFAR10~\citep{cifar} with $(\varepsilon,\delta)=(1, 10^{-5}$). The performance on the downstream task without privacy is $94.10\%$~\citep{cvpr}. 
We optimize the model with a learning rate $\classifierlr=0.1$ for parameters of classification layer (which is randomly intialized) and a learning rate $\lr=0.01$ for the remaining parameters. The mask finding epoch $T_0 = 5$ for the ablation study and the remaining hyperparameters are similar to the ones introduced in \cref{sec:experiments} ($T=50$, $C=1$, $\text{batch\_size}=500$, etc.)
We use 5 independent runs for the error bars in the ablation study. Specifically, we consider the following alternatives: \\
(1) ~\averagedtrue where we use true (unclipped and not noisy) gradients averaged over an epoch to solve Problem~\eqref{diagonal-upper-norm1} without any grouping. Training is still done with DP-SGD. We show the details of this procedure in Algorithm~\ref{algo:ex2}. This procedure is not $(\varepsilon,\delta)$-DP, but allows us to characterize the best possible performance of DP-SGD, if we had infinite privacy budget for choosing the mask using our optimization formulation introduced in \cref{general-1}. \\
(2) ~\averagednoisy which is mask selection procedure similar to ~\averagedtrue, but instead of using true gradients, we use those available from DP-SGD to ensure that the end-to-end procedure satisfies DP. This is the same procedure as in Example~\ref{example1}, and is shown in Algorithm~\ref{algo:ex1}.\\

\noindent We show the results for these methods in Table~\ref{tab:ablation_resnet18}. We see that ~\averagedtrue has the best accuracy but this procedure does not satisfy DP. However, we can deduce that the best accuracy we can expect from DP-SGD with a good mask is $79.26\%$. As soon as we use DP-SGD gradients for mask selection in ~\averagednoisy, we see a significant drop in accuracy which shows that a simple implementation of mask selection using DP-SGD gradients does not lead to significant improvements. On the other hand, ~\ourmethod leads to a better performance compared to ~\averagednoisy. This shows the success of our mask selection procedure that makes use of grouping, under DP constraints. We opt for a ~\rowgrouping for simplicity and the hardware speedups enabled by this group structure (see \cref{main:sec-hardware}).
\newline
\textbf{Ablation study conclusion:} The ablation study shows that our optimization formulation in Problem~\eqref{diagonal-upper-norm1} is a good direction towards improving the utility of DP-SGD training by leveraging sparse finetuning. Naively using privatized gradients to solve this problem is not the correct idea given the high-privacy regime which makes ranking a large number of data points with too much noise a hard task. Instead, we propose grouping these data points to increase the signal-to-noise ratio. This allows to rank less number of data points with less noise variance. Our approach is robust to the choice of grouping and ~\rowgrouping is a heuristic that works well in practice and can result in speedups with an improved implementation.
\begin{table}[!h]
\footnotesize
    \caption{Ablation study for ResNet18 on the DP finetuning task CIFAR100/CIFAR10 with $(\varepsilon,\delta) = (1, 10^{-5})$ that led to the development of \ourmethod.}
    \label{tab:ablation_resnet18}
\centering
\begin{tabular}{cccc}
  \toprule
  Method Name & Grouping & Accuracy \\
  \midrule
  \averagedtrue & \ding{55} & $79.26\pm(0.14)\%$\\
  \averagednoisy & \ding{55} & $77.89\pm(0.05)\%$\\
  \texttt{Random-Groups} & \ding{55} & $77.84\pm(0.17)\%$\\
  \ourmethod & \ding{51} & $78.81\pm(0.12)\%$\\
  \bottomrule
\end{tabular}
\end{table}

\begin{figure}
    \centering
    \includegraphics[width=0.455\textwidth]{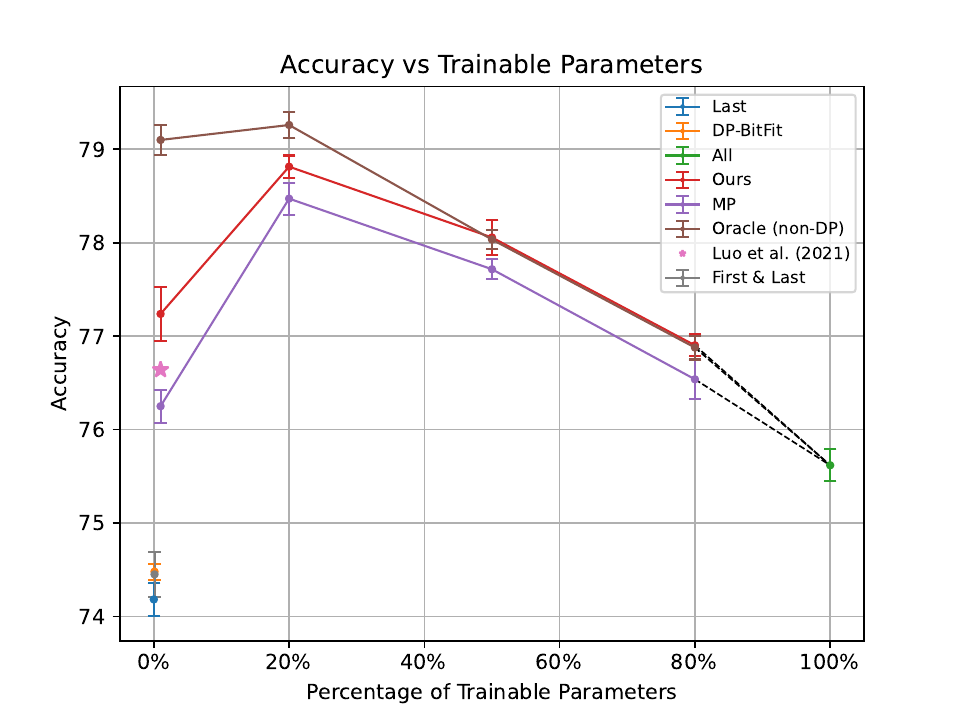}
    \caption{Profile of Accuracy/Percentage of trainable parameters for ResNet18 under $(\varepsilon, \delta) = (1, 10^{-5})$ DP-guarantees.}
    \label{fig:resnet_interpolation}
\end{figure}

To further understand the role of sparsity, we profile the performance of ~\ourmethod with different sparsity patterns and compare with other sparse fine-tuning methods (some of which are introduced in \cref{sec:experiments}) under the same setting (hyperparameter choices and privacy guarantees) in \cref{fig:resnet_interpolation}.

\section{Experiments}
\label{sec:experiments}
In this section, we study the performance of ~\ourmethod and compare with the state of the art methods for sparse fine-tuning. The implementation is done with Opacus \cite{opacus} with default settings (privacy accountant, batch sampling..).\\
\noindent\textbf{Datasets and models:} We explore several computer vision architectures including vision transformers (in particular DeiT) and Wide-ResNet. In terms of vision transformers, we study tiny, small, and base variants of DeiTs~\citep{deit}, pre-trained on ImageNet~\citep{imagenet}. Additionally, we consider the Wide-ResNet architecture WRN-28-10 pretrained on ImageNet-32.
The considered fine-tuning tasks are CIFAR10/100 \citep{cifar}.\\
\noindent\textbf{Competing methods:} We consider full fine-tuning (All), last layer (Last) fine-tuning following~\citep{deepmindft}, a mask selection based on the magnitude of public pre-trained weights \mpweights, as introduced in \cref{algo:mpweights-alg}, and DP-BitFit ~\citep{bu2022differentially}. 
We follow the same hyper-parameter tuning procedure discussed in Appendix~\ref{app:tuning} for ~\ourmethod, as well as the competing methods. The entire last layer and group normalization parameters are trainable parameters. This minimal number corresponds to (Last). Adding bias terms to the trainable set corresponds to (DP-BitFit). For ~\ourmethod and ~\mpweights, the last layer, group normalization and bias terms are trainable; and the mask selection happens on the convolution/linear parameters.

\noindent\textbf{Details on Hyper-parameters:}\label{app:tuning} 
In all of our experiments, we use a fixed $\delta=10^{-5}$ and varying values of $\varepsilon \in (2, 4, 8)$ for our $(\varepsilon, \delta)$ guarantees corresponding to a high to moderate privacy regime. We fix the percentage of trainable parameters in ~\ourmethod to $20\%$, the number of epochs $T=50$, the mask finding epoch $T_0 = 10$, the clipping constant $C = 1$ and the $\text{batch\_size}=500$. We use SGD as optimizer with fixed $\text{momentum}=0.9$ and no weight decay. We use a cosine annealing scheduler in all experiments with a linear $\text{warm\_up}=0.02$ -- corresponding to one epoch. All the models used have been pre-trained on their corresponding public datasets with group normalization. For the transfer learning task, we randomly initialize the Last Layer since the output dimension changes. The learning rate $\lr=0.01$ for all experiments.
\newline 
For the mask selection procedure based on the magnitude of the pre-trained weights as introduced in \citet{cvpr}, we use a percentage of trainable parameters equal to $20\%$ (similar to the percentage of trainable parameters reported for ~\ourmethod) instead of the reported $p=1\%$ because it gives a better performance on our considered model/dataset/privacy combinations.

\noindent\textbf{Results:} \cref{tab:table_results} shows that ~\ourmethod performs better than all layers fine-tuning, confirming that sparse fine-tuning under DP can improve the prediction accuracy. We also see that ~\ourmethod performs better than existing sparse DP fine-tuning methods, which shows the benefits of our optimization-based approach to sparse subnetwork selection.

\begin{table*}[h!]
\footnotesize
\centering
\resizebox{0.92\textwidth}{!}{%
\renewcommand{\arraystretch}{1.3}
\begin{tabular}{ccccccccc}
  \toprule
  Privacy & Dataset & Model & \ourmethod & All & Last & \mpweights & DP-BitFit~\cite{bu2022differentially} & Non-Private\\
  \midrule
  \multirow{8.25}{*}{$\epsilon = 2$}
  &\multirow{4}{*}{CIFAR10} 
  & DeiT Tiny & \textbf{92.82}$\pm$(0.06)\% & 90.38$\pm$(0.13)\% & 85.65$\pm$(0.02)\% &  92.69$\pm$(0.03)\%& 92.24$\pm$(0.04)\% &97.20\%\\
  &&DeiT Small &\textbf{95.71}$\pm$(0.04)\% & 94.29$\pm$(0.10)\% & 90.13$\pm$(0.05)\% &  95.55$\pm$(0.04)\%& 95.19$\pm$(0.03)\%& 98.14\%\\
  &&DeiT Base &\textbf{96.80}$\pm$(0.03)\% & 96.28$\pm$(0.09)\% & 92.38$\pm$(0.05)\% &  96.70$\pm$(0.05)\%& 96.29$\pm$(0.02)\% & 98.53\%\\
  &&WRN-2810 &\textbf{96.42}$\pm$(0.04)\% &95.63$\pm$(0.05)\% &96.26$\pm$(0.03)\% &96.30$\pm$(0.05)\% &96.26$\pm$(0.04)\% &97.76\%\\
  \cmidrule(rl){2-9}
  &\multirow{4}{*}{CIFAR100} 
  & DeiT Tiny & \textbf{69.58}$\pm$(0.10)\% & 61.22$\pm$(0.36)\%& 58.10$\pm$(0.09)\%&  69.01$\pm$(0.25)\%& 69.53$\pm$(0.13)\%&84.86\%\\
  &&DeiT Small& \textbf{75.87}$\pm$(0.21)\%& 71.21$\pm$(0.22)\% & 64.57$\pm$(0.07)\% &  75.32$\pm$(0.05)\%&{75.26}$\pm$(0.16)\%&88.15\%\\
  &&DeiT Base& \textbf{79.91}$\pm$(0.11)\%& 78.232$\pm$(0.12)\% & 66.44$\pm$(0.11)\% &  79.64$\pm$(0.10)\% & 78.24$\pm$(0.08)\% & 90.53\%\\
  &&WRN-2810 &\textbf{73.59}$\pm$(0.09)\% &70.16$\pm$(0.20)\% &73.40$\pm$(0.07)\% &73.10$\pm$(0.11)\% &72.84$\pm$(0.13)\% &85.11\%\\

  \midrule
  \multirow{8.25}{*}{$\epsilon = 4$}
  &\multirow{4}{*}{CIFAR10} 
  &Deit Tiny &\textbf{93.33}$\pm$(0.03)\% &92.30$\pm$(0.15)\% &85.70$\pm$(0.02)\% &93.15$\pm$(0.02)\% &92.25$\pm$(0.04)\% &97.20\% \\
  &&Deit Small &\textbf{95.96}$\pm$(0.04)\% &95.53$\pm$(0.06)\% &90.15$\pm$(0.07)\% &95.78$\pm$(0.03)\% &95.23$\pm$(0.01)\% &98.14\% \\
  &&Deit Base &\textbf{96.98}$\pm$(0.03)\% &96.83$\pm$(0.07)\% &92.37$\pm$(0.05)\% &96.80$\pm$(0.05)\% &96.30$\pm$(0.02)\% &98.53\% \\
  &&WRN-2810 &\textbf{96.65}$\pm$(0.02)\% &96.21$\pm$(0.03)\% &96.31$\pm$(0.04)\% &96.44$\pm$(0.05)\% &96.32$\pm$(0.04)\% &97.76\% \\
  \cmidrule(rl){2-9}
  &\multirow{4}{*}{CIFAR100} 
  &Deit Tiny &\textbf{70.22}$\pm$(0.13)\% &64.74$\pm$(0.21)\% &58.23$\pm$(0.10)\% &69.91$\pm$(0.20)\% &69.73$\pm$(0.10)\% &84.86\% \\  
  &&Deit Small &\textbf{76.51}$\pm$(0.16)\% &73.99$\pm$(0.21)\% &64.64$\pm$(0.05)\% &76.21$\pm$(0.16)\% &75.31$\pm$(0.11)\% &88.15\% \\
  &&Deit Base &\textbf{80.37}$\pm$(0.09)\% &79.69$\pm$(0.09)\% &66.57$\pm$(0.09)\% &80.03$\pm$(0.11)\% &78.38$\pm$(0.08)\% &90.53\% \\
  &&WRN-2810 &\textbf{74.24}$\pm$(0.11)\% &72.46$\pm$(0.21)\% &73.62$\pm$(0.08)\% &73.85$\pm$(0.07)\% &73.19$\pm$(0.09)\% &85.11\% \\
  
  \midrule
  \multirow{8.25}{*}{$\epsilon = 8$}
  &\multirow{4}{*}{CIFAR10} 
  &Deit Tiny &\textbf{93.75}$\pm$(0.05)\% &92.92$\pm$(0.15)\% &85.71$\pm$(0.03)\% &93.27$\pm$(0.04)\% &92.27$\pm$(0.05)\% &97.20\%\\
  &&Deit Small &\textbf{96.12}$\pm$(0.07)\% &95.89$\pm$(0.07)\% &90.17$\pm$(0.07)\% &95.85$\pm$(0.02)\% &95.25$\pm$(0.02)\% &98.14\% \\
  &&Deit Base &\textbf{97.05}$\pm$(0.03)\% &97.99$\pm$(0.05)\% &92.33$\pm$(0.07)\% &96.85$\pm$(0.03)\% &96.30$\pm$(0.03)\% &98.53\% \\
  &&WRN-2810 &\textbf{96.70}$\pm$(0.02)\% &96.33$\pm$(0.03)\% &96.33$\pm$(0.03)\% &96.49$\pm$(0.06)\% &96.33$\pm$(0.04)\% &97.76\% \\
  \cmidrule(rl){2-9}
  &\multirow{4}{*}{CIFAR100} 
  &Deit Tiny &\textbf{70.54}$\pm$(0.14)\% &65.91$\pm$(0.23)\% &58.27$\pm$(0.08)\% &70.20$\pm$(0.20)\% &69.82$\pm$(0.11)\% &84.86\% \\  
  &&Deit Small &\textbf{76.72}$\pm$(0.17)\% &74.71$\pm$(0.18)\% &64.66$\pm$(0.05)\% &76.49$\pm$(0.14)\% &75.37$\pm$(0.11)\% &88.15\% \\
  &&Deit Base &\textbf{80.40}$\pm$(0.10)\% &80.12$\pm$(0.07)\% &66.69$\pm$(0.10)\% &80.11$\pm$(0.10)\% &78.51$\pm$(0.08)\% &90.53\% \\
  &&WRN-2810 &\textbf{74.49}$\pm$(0.17)\% &73.11$\pm$(0.21)\% &73.70$\pm$(0.13)\% &74.28$\pm$(0.09)\% &73.42$\pm$(0.16)\% &85.11\% \\
  \bottomrule
\end{tabular}
}
\caption{Numerical results for fine-tuning vision models with varying $\epsilon$ values and a fixed $\delta=10^{-5}$ $(\varepsilon, \delta)$ DP-guarantees for different sparse fine-tuning methods.}
\label{tab:table_results}
\end{table*}

\section{Improved Implementation for Hardware speedups}
\label{main:sec-hardware}

\begin{figure}
    \centering
    \includegraphics[width=0.5\textwidth]{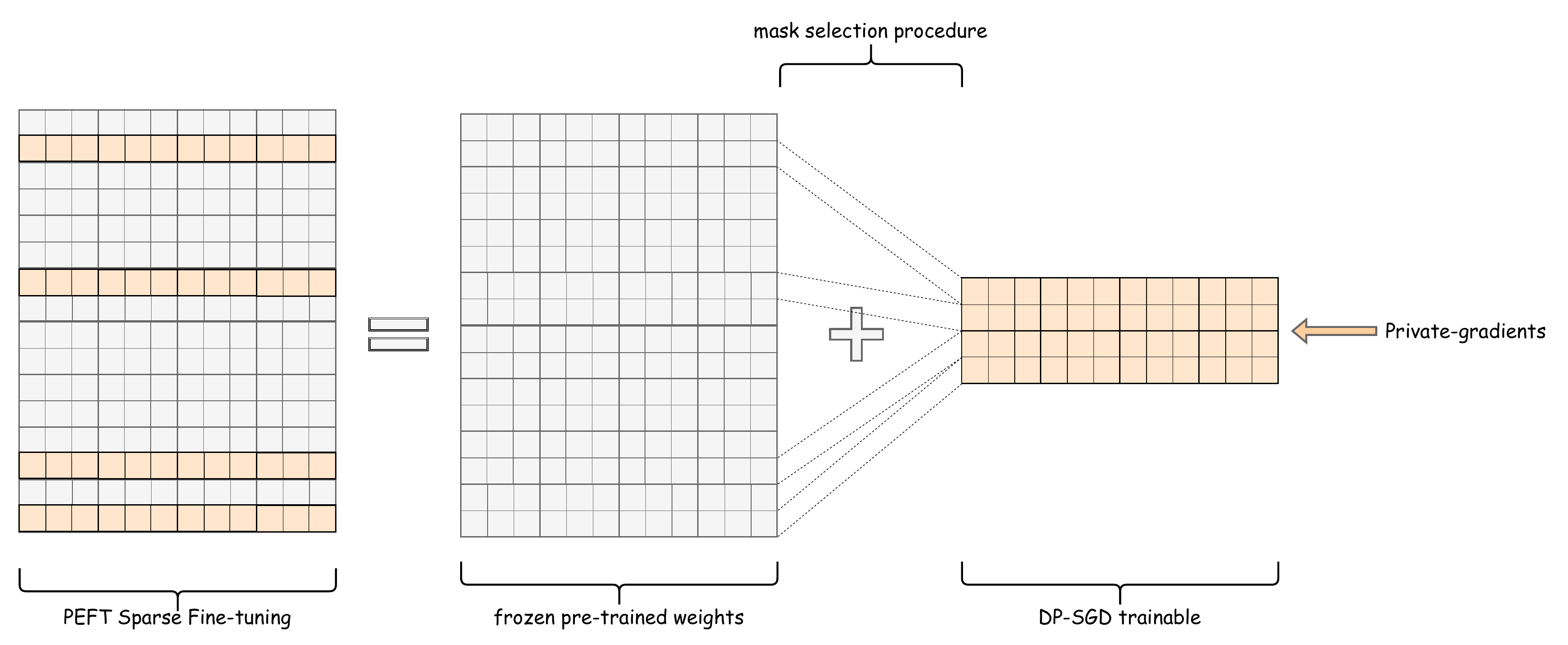}
    \caption{Efficient implementation of sparse DP-SGD fine-tuning of our proposed \rowgrouping scheme.}
    \label{fig:sft_peft_pdf}
\end{figure}

Our introduced \rowgrouping scheme yields a structure in terms of fine-tuning that can be exploited for hardware speedups. We propose an improved implementation where the trainable groups are stacked together as a separate matrix that receives private gradient updates, whereas the initial pre-trained weights $\bf W_\text{old}$ remain fixed. For instance, the introduced implementation as shown in \cref{fig:sft_peft_pdf} results in a $15\%$ wall-clock time speedup compared to standard full fine-tuning for the DeiT Base model using sparse fine-tuning with the number of trainable parameters set to $1\%$---benchmarked on an L40 and A100 GPU.

\section{Discussion and Limitations}
\label{main:sec-discussion-limitations}

The numerical experiments support that sparse fine-tuning can be valuable in DP, and that algorithmically finding a mask that results in a performance boost is achievable thanks to \ourmethod. The framework we propose comes with a few limitations. Even though we propose default values for $T_0 = 10$ and the sparsity pattern $k/q = 20\%$ used in the experiments \cref{sec:experiments}, the optimal way to find these values is to perform a hyperparameter search. That being said, \ourmethod seems to be robust to changes in these hyperparameters; \ourmethod is a \textit{smart interpolation} between All layers finetuning (\ourmethod reduces to All layer fine-tuning if $k/q = 100\%$) and DP-BitFit (\ourmethod reduces to DP-BitFit if $k/q = 0\%$). The hyperparameters control whether \ourmethod performs like All layers fine-tuning  or DP-BitFit fine-tuning (see \cref{fig:resnet_interpolation} and \cref{fig:tiny_interpolation}).

We believe that our findings can be further used to improve the utility of other PEFT methods which introduce additional trainable parameters like LoRA~\cite{hu2022lora} for DP-learning. An adaptation to our optimization formulation \eqref{general-1} can be introduced for rank-selection in DP-LoRA~\cite{liu2023differentially}. A similar subspace selection procedure (as opposed to row-group selection in this paper) can be applied to select a LoRA rank without expensive hyperparameter tuning.

\section{Conclusion}
\label{main:sec-conclusion}

We explore the effect of sparse fine-tuning on DP-SGD dynamics and present a simple and portable mask selection procedure that could be plugged on top of standard DP-SGD (and DP-SGD like algorithms) in neural network private fine-tuning to improve the utility of vanilla DP-SGD. The introduced structure in sparse fine-tuning can further be exploited to obtain hardware speedups
compared to standard DP-SGD full fine-tuning of the model.

\vspace{-5pt}
\section{Acknowledgements}
\vspace{-3pt}
This research is supported in part by grants from Google and the Office of Naval Research. We acknowledge MIT SuperCloud~\cite{reuther2018interactive} for providing HPC resources that have contributed to the research results reported within this paper. We also acknowledge Google Cloud Credits for computing. 
Kayhan and Hussein contributed to this research while at MIT and Google, respectively.

\clearpage
\newpage
\bibliography{dpruning}


\begin{thebibliography}{40}


\ifx \showCODEN    \undefined \def \showCODEN     #1{\unskip}     \fi
\ifx \showDOI      \undefined \def \showDOI       #1{#1}\fi
\ifx \showISBNx    \undefined \def \showISBNx     #1{\unskip}     \fi
\ifx \showISBNxiii \undefined \def \showISBNxiii  #1{\unskip}     \fi
\ifx \showISSN     \undefined \def \showISSN      #1{\unskip}     \fi
\ifx \showLCCN     \undefined \def \showLCCN      #1{\unskip}     \fi
\ifx \shownote     \undefined \def \shownote      #1{#1}          \fi
\ifx \showarticletitle \undefined \def \showarticletitle #1{#1}   \fi
\ifx \showURL      \undefined \def \showURL       {\relax}        \fi
\providecommand\bibfield[2]{#2}
\providecommand\bibinfo[2]{#2}
\providecommand\natexlab[1]{#1}
\providecommand\showeprint[2][]{arXiv:#2}

\bibitem[Abadi et~al\mbox{.}(2016)]%
        {abadi2016deep}
\bibfield{author}{\bibinfo{person}{Martin Abadi}, \bibinfo{person}{Andy Chu}, \bibinfo{person}{Ian Goodfellow}, \bibinfo{person}{H~Brendan McMahan}, \bibinfo{person}{Ilya Mironov}, \bibinfo{person}{Kunal Talwar}, {and} \bibinfo{person}{Li Zhang}.} \bibinfo{year}{2016}\natexlab{}.
\newblock \showarticletitle{Deep learning with differential privacy}. In \bibinfo{booktitle}{\emph{Proceedings of the 2016 ACM SIGSAC conference on computer and communications security}}. \bibinfo{pages}{308--318}.
\newblock


\bibitem[Adamczewski et~al\mbox{.}(2023)]%
        {preprune}
\bibfield{author}{\bibinfo{person}{Kamil Adamczewski}, \bibinfo{person}{Yingchen He}, {and} \bibinfo{person}{Mijung Park}.} \bibinfo{year}{2023}\natexlab{}.
\newblock \showarticletitle{Pre-Pruning and Gradient-Dropping Improve Differentially Private Image Classification}.
\newblock \bibinfo{journal}{\emph{arXiv preprint arXiv:2306.11754}} (\bibinfo{year}{2023}).
\newblock


\bibitem[Bassily et~al\mbox{.}(2014)]%
        {normd1}
\bibfield{author}{\bibinfo{person}{Raef Bassily}, \bibinfo{person}{Adam Smith}, {and} \bibinfo{person}{Abhradeep Thakurta}.} \bibinfo{year}{2014}\natexlab{}.
\newblock \showarticletitle{Private empirical risk minimization: Efficient algorithms and tight error bounds}. In \bibinfo{booktitle}{\emph{2014 IEEE 55th annual symposium on foundations of computer science}}. IEEE, \bibinfo{pages}{464--473}.
\newblock


\bibitem[Benbaki et~al\mbox{.}(2023)]%
        {chita}
\bibfield{author}{\bibinfo{person}{Riade Benbaki}, \bibinfo{person}{Wenyu Chen}, \bibinfo{person}{Xiang Meng}, \bibinfo{person}{Hussein Hazimeh}, \bibinfo{person}{Natalia Ponomareva}, \bibinfo{person}{Zhe Zhao}, {and} \bibinfo{person}{Rahul Mazumder}.} \bibinfo{year}{2023}\natexlab{}.
\newblock \showarticletitle{Fast as chita: Neural network pruning with combinatorial optimization}. In \bibinfo{booktitle}{\emph{International Conference on Machine Learning}}. PMLR, \bibinfo{pages}{2031--2049}.
\newblock


\bibitem[Bu et~al\mbox{.}(2022)]%
        {bu2022differentially}
\bibfield{author}{\bibinfo{person}{Zhiqi Bu}, \bibinfo{person}{Yu-Xiang Wang}, \bibinfo{person}{Sheng Zha}, {and} \bibinfo{person}{George Karypis}.} \bibinfo{year}{2022}\natexlab{}.
\newblock \showarticletitle{Differentially private bias-term only fine-tuning of foundation models}.
\newblock  (\bibinfo{year}{2022}).
\newblock


\bibitem[Cattan et~al\mbox{.}(2022)]%
        {firstlastwrn}
\bibfield{author}{\bibinfo{person}{Yannis Cattan}, \bibinfo{person}{Christopher~A Choquette-Choo}, \bibinfo{person}{Nicolas Papernot}, {and} \bibinfo{person}{Abhradeep Thakurta}.} \bibinfo{year}{2022}\natexlab{}.
\newblock \showarticletitle{Fine-tuning with differential privacy necessitates an additional hyperparameter search}.
\newblock \bibinfo{journal}{\emph{arXiv preprint arXiv:2210.02156}} (\bibinfo{year}{2022}).
\newblock


\bibitem[Choquette-Choo et~al\mbox{.}(2024)]%
        {choquette2024amplified}
\bibfield{author}{\bibinfo{person}{Christopher~A Choquette-Choo}, \bibinfo{person}{Arun Ganesh}, \bibinfo{person}{Ryan McKenna}, \bibinfo{person}{H~Brendan McMahan}, \bibinfo{person}{John Rush}, \bibinfo{person}{Abhradeep Guha~Thakurta}, {and} \bibinfo{person}{Zheng Xu}.} \bibinfo{year}{2024}\natexlab{}.
\newblock \showarticletitle{(Amplified) Banded Matrix Factorization: A unified approach to private training}.
\newblock \bibinfo{journal}{\emph{Advances in Neural Information Processing Systems}}  \bibinfo{volume}{36} (\bibinfo{year}{2024}).
\newblock


\bibitem[De et~al\mbox{.}(2022)]%
        {deepmindft}
\bibfield{author}{\bibinfo{person}{Soham De}, \bibinfo{person}{Leonard Berrada}, \bibinfo{person}{Jamie Hayes}, \bibinfo{person}{Samuel~L Smith}, {and} \bibinfo{person}{Borja Balle}.} \bibinfo{year}{2022}\natexlab{}.
\newblock \showarticletitle{Unlocking high-accuracy differentially private image classification through scale}.
\newblock \bibinfo{journal}{\emph{arXiv preprint arXiv:2204.13650}} (\bibinfo{year}{2022}).
\newblock


\bibitem[Deng et~al\mbox{.}(2009)]%
        {imagenet}
\bibfield{author}{\bibinfo{person}{Jia Deng}, \bibinfo{person}{Wei Dong}, \bibinfo{person}{Richard Socher}, \bibinfo{person}{Li-Jia Li}, \bibinfo{person}{Kai Li}, {and} \bibinfo{person}{Li Fei-Fei}.} \bibinfo{year}{2009}\natexlab{}.
\newblock \showarticletitle{Imagenet: A large-scale hierarchical image database}. In \bibinfo{booktitle}{\emph{2009 IEEE conference on computer vision and pattern recognition}}. Ieee, \bibinfo{pages}{248--255}.
\newblock


\bibitem[Dwork(2006)]%
        {dwork2006differential}
\bibfield{author}{\bibinfo{person}{Cynthia Dwork}.} \bibinfo{year}{2006}\natexlab{}.
\newblock \showarticletitle{Differential privacy}. In \bibinfo{booktitle}{\emph{International colloquium on automata, languages, and programming}}. Springer, \bibinfo{pages}{1--12}.
\newblock


\bibitem[Dwork et~al\mbox{.}(2014)]%
        {dwork2014algorithmic}
\bibfield{author}{\bibinfo{person}{Cynthia Dwork}, \bibinfo{person}{Aaron Roth}, {et~al\mbox{.}}} \bibinfo{year}{2014}\natexlab{}.
\newblock \showarticletitle{The algorithmic foundations of differential privacy}.
\newblock \bibinfo{journal}{\emph{Foundations and Trends{\textregistered} in Theoretical Computer Science}} \bibinfo{volume}{9}, \bibinfo{number}{3--4} (\bibinfo{year}{2014}), \bibinfo{pages}{211--407}.
\newblock


\bibitem[Frantar and Alistarh(2022)]%
        {obc}
\bibfield{author}{\bibinfo{person}{Elias Frantar} {and} \bibinfo{person}{Dan Alistarh}.} \bibinfo{year}{2022}\natexlab{}.
\newblock \showarticletitle{Optimal brain compression: A framework for accurate post-training quantization and pruning}.
\newblock \bibinfo{journal}{\emph{Advances in Neural Information Processing Systems}}  \bibinfo{volume}{35} (\bibinfo{year}{2022}), \bibinfo{pages}{4475--4488}.
\newblock


\bibitem[Ganesh et~al\mbox{.}(2023)]%
        {ganesh2023public}
\bibfield{author}{\bibinfo{person}{Arun Ganesh}, \bibinfo{person}{Mahdi Haghifam}, \bibinfo{person}{Milad Nasr}, \bibinfo{person}{Sewoong Oh}, \bibinfo{person}{Thomas Steinke}, \bibinfo{person}{Om Thakkar}, \bibinfo{person}{Abhradeep~Guha Thakurta}, {and} \bibinfo{person}{Lun Wang}.} \bibinfo{year}{2023}\natexlab{}.
\newblock \showarticletitle{Why is public pretraining necessary for private model training?}. In \bibinfo{booktitle}{\emph{International Conference on Machine Learning}}. PMLR, \bibinfo{pages}{10611--10627}.
\newblock


\bibitem[Gopi et~al\mbox{.}(2021)]%
        {prv-opacus}
\bibfield{author}{\bibinfo{person}{Sivakanth Gopi}, \bibinfo{person}{Yin~Tat Lee}, {and} \bibinfo{person}{Lukas Wutschitz}.} \bibinfo{year}{2021}\natexlab{}.
\newblock \showarticletitle{Numerical composition of differential privacy}.
\newblock \bibinfo{journal}{\emph{Advances in Neural Information Processing Systems}}  \bibinfo{volume}{34} (\bibinfo{year}{2021}), \bibinfo{pages}{11631--11642}.
\newblock


\bibitem[Hassibi et~al\mbox{.}(1993)]%
        {obs}
\bibfield{author}{\bibinfo{person}{Babak Hassibi}, \bibinfo{person}{David~G Stork}, {and} \bibinfo{person}{Gregory~J Wolff}.} \bibinfo{year}{1993}\natexlab{}.
\newblock \showarticletitle{Optimal brain surgeon and general network pruning}. In \bibinfo{booktitle}{\emph{IEEE international conference on neural networks}}. IEEE, \bibinfo{pages}{293--299}.
\newblock


\bibitem[He et~al\mbox{.}(2016)]%
        {resnet}
\bibfield{author}{\bibinfo{person}{Kaiming He}, \bibinfo{person}{Xiangyu Zhang}, \bibinfo{person}{Shaoqing Ren}, {and} \bibinfo{person}{Jian Sun}.} \bibinfo{year}{2016}\natexlab{}.
\newblock \showarticletitle{Deep residual learning for image recognition}. In \bibinfo{booktitle}{\emph{Proceedings of the IEEE conference on computer vision and pattern recognition}}. \bibinfo{pages}{770--778}.
\newblock


\bibitem[He et~al\mbox{.}(2021)]%
        {he2021effectiveness}
\bibfield{author}{\bibinfo{person}{Ruidan He}, \bibinfo{person}{Linlin Liu}, \bibinfo{person}{Hai Ye}, \bibinfo{person}{Qingyu Tan}, \bibinfo{person}{Bosheng Ding}, \bibinfo{person}{Liying Cheng}, \bibinfo{person}{Jia-Wei Low}, \bibinfo{person}{Lidong Bing}, {and} \bibinfo{person}{Luo Si}.} \bibinfo{year}{2021}\natexlab{}.
\newblock \showarticletitle{On the effectiveness of adapter-based tuning for pretrained language model adaptation}.
\newblock \bibinfo{journal}{\emph{arXiv preprint arXiv:2106.03164}} (\bibinfo{year}{2021}).
\newblock


\bibitem[Hu et~al\mbox{.}(2022)]%
        {hu2022lora}
\bibfield{author}{\bibinfo{person}{Edward~J Hu}, \bibinfo{person}{yelong shen}, \bibinfo{person}{Phillip Wallis}, \bibinfo{person}{Zeyuan Allen-Zhu}, \bibinfo{person}{Yuanzhi Li}, \bibinfo{person}{Shean Wang}, \bibinfo{person}{Lu Wang}, {and} \bibinfo{person}{Weizhu Chen}.} \bibinfo{year}{2022}\natexlab{}.
\newblock \showarticletitle{Lo{RA}: Low-Rank Adaptation of Large Language Models}. In \bibinfo{booktitle}{\emph{International Conference on Learning Representations}}.
\newblock
\urldef\tempurl%
\url{https://openreview.net/forum?id=nZeVKeeFYf9}
\showURL{%
\tempurl}


\bibitem[Jia et~al\mbox{.}(2022)]%
        {jia2022visual}
\bibfield{author}{\bibinfo{person}{Menglin Jia}, \bibinfo{person}{Luming Tang}, \bibinfo{person}{Bor-Chun Chen}, \bibinfo{person}{Claire Cardie}, \bibinfo{person}{Serge Belongie}, \bibinfo{person}{Bharath Hariharan}, {and} \bibinfo{person}{Ser-Nam Lim}.} \bibinfo{year}{2022}\natexlab{}.
\newblock \showarticletitle{Visual prompt tuning}. In \bibinfo{booktitle}{\emph{European Conference on Computer Vision}}. Springer, \bibinfo{pages}{709--727}.
\newblock


\bibitem[Krizhevsky et~al\mbox{.}(2009)]%
        {cifar}
\bibfield{author}{\bibinfo{person}{Alex Krizhevsky}, \bibinfo{person}{Geoffrey Hinton}, {et~al\mbox{.}}} \bibinfo{year}{2009}\natexlab{}.
\newblock \showarticletitle{Learning multiple layers of features from tiny images}.
\newblock  (\bibinfo{year}{2009}).
\newblock


\bibitem[Kurakin et~al\mbox{.}(2022)]%
        {kurakin2022toward}
\bibfield{author}{\bibinfo{person}{Alexey Kurakin}, \bibinfo{person}{Shuang Song}, \bibinfo{person}{Steve Chien}, \bibinfo{person}{Roxana Geambasu}, \bibinfo{person}{Andreas Terzis}, {and} \bibinfo{person}{Abhradeep Thakurta}.} \bibinfo{year}{2022}\natexlab{}.
\newblock \showarticletitle{Toward training at imagenet scale with differential privacy}.
\newblock \bibinfo{journal}{\emph{arXiv preprint arXiv:2201.12328}} (\bibinfo{year}{2022}).
\newblock


\bibitem[LeCun et~al\mbox{.}(1989)]%
        {obd}
\bibfield{author}{\bibinfo{person}{Yann LeCun}, \bibinfo{person}{John Denker}, {and} \bibinfo{person}{Sara Solla}.} \bibinfo{year}{1989}\natexlab{}.
\newblock \showarticletitle{Optimal brain damage}.
\newblock \bibinfo{journal}{\emph{Advances in neural information processing systems}}  \bibinfo{volume}{2} (\bibinfo{year}{1989}).
\newblock


\bibitem[Li et~al\mbox{.}(2022)]%
        {li2022does}
\bibfield{author}{\bibinfo{person}{Xuechen Li}, \bibinfo{person}{Daogao Liu}, \bibinfo{person}{Tatsunori~B Hashimoto}, \bibinfo{person}{Huseyin~A Inan}, \bibinfo{person}{Janardhan Kulkarni}, \bibinfo{person}{Yin-Tat Lee}, {and} \bibinfo{person}{Abhradeep Guha~Thakurta}.} \bibinfo{year}{2022}\natexlab{}.
\newblock \showarticletitle{When does differentially private learning not suffer in high dimensions?}
\newblock \bibinfo{journal}{\emph{Advances in Neural Information Processing Systems}}  \bibinfo{volume}{35} (\bibinfo{year}{2022}), \bibinfo{pages}{28616--28630}.
\newblock


\bibitem[Li et~al\mbox{.}(2021)]%
        {dplora2}
\bibfield{author}{\bibinfo{person}{Xuechen Li}, \bibinfo{person}{Florian Tramer}, \bibinfo{person}{Percy Liang}, {and} \bibinfo{person}{Tatsunori Hashimoto}.} \bibinfo{year}{2021}\natexlab{}.
\newblock \showarticletitle{Large language models can be strong differentially private learners}.
\newblock \bibinfo{journal}{\emph{arXiv preprint arXiv:2110.05679}} (\bibinfo{year}{2021}).
\newblock


\bibitem[Liu et~al\mbox{.}(2023)]%
        {liu2023differentially}
\bibfield{author}{\bibinfo{person}{Xiao-Yang Liu}, \bibinfo{person}{Rongyi Zhu}, \bibinfo{person}{Daochen Zha}, \bibinfo{person}{Jiechao Gao}, \bibinfo{person}{Shan Zhong}, \bibinfo{person}{Matt White}, {and} \bibinfo{person}{Meikang Qiu}.} \bibinfo{year}{2023}\natexlab{}.
\newblock \showarticletitle{Differentially private low-rank adaptation of large language model using federated learning}.
\newblock \bibinfo{journal}{\emph{ACM Transactions on Management Information Systems}} (\bibinfo{year}{2023}).
\newblock


\bibitem[Luo et~al\mbox{.}(2021)]%
        {cvpr}
\bibfield{author}{\bibinfo{person}{Zelun Luo}, \bibinfo{person}{Daniel~J Wu}, \bibinfo{person}{Ehsan Adeli}, {and} \bibinfo{person}{Li Fei-Fei}.} \bibinfo{year}{2021}\natexlab{}.
\newblock \showarticletitle{Scalable differential privacy with sparse network finetuning}. In \bibinfo{booktitle}{\emph{Proceedings of the IEEE/CVF Conference on Computer Vision and Pattern Recognition}}. \bibinfo{pages}{5059--5068}.
\newblock


\bibitem[Machanavajjhala et~al\mbox{.}(2007)]%
        {privacy1}
\bibfield{author}{\bibinfo{person}{Ashwin Machanavajjhala}, \bibinfo{person}{Daniel Kifer}, \bibinfo{person}{Johannes Gehrke}, {and} \bibinfo{person}{Muthuramakrishnan Venkitasubramaniam}.} \bibinfo{year}{2007}\natexlab{}.
\newblock \showarticletitle{l-diversity: Privacy beyond k-anonymity}.
\newblock \bibinfo{journal}{\emph{Acm transactions on knowledge discovery from data (tkdd)}} \bibinfo{volume}{1}, \bibinfo{number}{1} (\bibinfo{year}{2007}), \bibinfo{pages}{3--es}.
\newblock


\bibitem[Mehta et~al\mbox{.}(2022)]%
        {mehta2022large}
\bibfield{author}{\bibinfo{person}{Harsh Mehta}, \bibinfo{person}{Abhradeep Thakurta}, \bibinfo{person}{Alexey Kurakin}, {and} \bibinfo{person}{Ashok Cutkosky}.} \bibinfo{year}{2022}\natexlab{}.
\newblock \showarticletitle{Large scale transfer learning for differentially private image classification}.
\newblock \bibinfo{journal}{\emph{arXiv preprint arXiv:2205.02973}} (\bibinfo{year}{2022}).
\newblock


\bibitem[Mironov et~al\mbox{.}(2019)]%
        {rdp1}
\bibfield{author}{\bibinfo{person}{Ilya Mironov}, \bibinfo{person}{Kunal Talwar}, {and} \bibinfo{person}{Li Zhang}.} \bibinfo{year}{2019}\natexlab{}.
\newblock \showarticletitle{R$\backslash$'enyi differential privacy of the sampled gaussian mechanism}.
\newblock \bibinfo{journal}{\emph{arXiv preprint arXiv:1908.10530}} (\bibinfo{year}{2019}).
\newblock


\bibitem[Papernot et~al\mbox{.}(2019)]%
        {papernot2019making}
\bibfield{author}{\bibinfo{person}{Nicolas Papernot}, \bibinfo{person}{Steve Chien}, \bibinfo{person}{Shuang Song}, \bibinfo{person}{Abhradeep Thakurta}, {and} \bibinfo{person}{Ulfar Erlingsson}.} \bibinfo{year}{2019}\natexlab{}.
\newblock \showarticletitle{Making the shoe fit: Architectures, initializations, and tuning for learning with privacy}.
\newblock  (\bibinfo{year}{2019}).
\newblock


\bibitem[Ponomareva et~al\mbox{.}(2023)]%
        {dpfy}
\bibfield{author}{\bibinfo{person}{Natalia Ponomareva}, \bibinfo{person}{Hussein Hazimeh}, \bibinfo{person}{Alex Kurakin}, \bibinfo{person}{Zheng Xu}, \bibinfo{person}{Carson Denison}, \bibinfo{person}{H.~Brendan McMahan}, \bibinfo{person}{Sergei Vassilvitskii}, \bibinfo{person}{Steve Chien}, {and} \bibinfo{person}{Abhradeep~Guha Thakurta}.} \bibinfo{year}{2023}\natexlab{}.
\newblock \showarticletitle{How to DP-fy {ML:} {A} Practical Guide to Machine Learning with Differential Privacy}.
\newblock \bibinfo{journal}{\emph{J. Artif. Intell. Res.}}  \bibinfo{volume}{77} (\bibinfo{year}{2023}), \bibinfo{pages}{1113--1201}.
\newblock
\urldef\tempurl%
\url{https://doi.org/10.1613/JAIR.1.14649}
\showDOI{\tempurl}


\bibitem[Reuther et~al\mbox{.}(2018)]%
        {reuther2018interactive}
\bibfield{author}{\bibinfo{person}{Albert Reuther}, \bibinfo{person}{Jeremy Kepner}, \bibinfo{person}{Chansup Byun}, \bibinfo{person}{Siddharth Samsi}, \bibinfo{person}{William Arcand}, \bibinfo{person}{David Bestor}, \bibinfo{person}{Bill Bergeron}, \bibinfo{person}{Vijay Gadepally}, \bibinfo{person}{Michael Houle}, \bibinfo{person}{Matthew Hubbell}, \bibinfo{person}{Michael Jones}, \bibinfo{person}{Anna Klein}, \bibinfo{person}{Lauren Milechin}, \bibinfo{person}{Julia Mullen}, \bibinfo{person}{Andrew Prout}, \bibinfo{person}{Antonio Rosa}, \bibinfo{person}{Charles Yee}, {and} \bibinfo{person}{Peter Michaleas}.} \bibinfo{year}{2018}\natexlab{}.
\newblock \showarticletitle{Interactive supercomputing on 40,000 cores for machine learning and data analysis}. In \bibinfo{booktitle}{\emph{2018 IEEE High Performance extreme Computing Conference (HPEC)}}. IEEE, \bibinfo{pages}{1–6}.
\newblock


\bibitem[Shen et~al\mbox{.}(2021)]%
        {dpsize}
\bibfield{author}{\bibinfo{person}{Yinchen Shen}, \bibinfo{person}{Zhiguo Wang}, \bibinfo{person}{Ruoyu Sun}, {and} \bibinfo{person}{Xiaojing Shen}.} \bibinfo{year}{2021}\natexlab{}.
\newblock \showarticletitle{Towards understanding the impact of model size on differential private classification}.
\newblock \bibinfo{journal}{\emph{arXiv preprint arXiv:2111.13895}} (\bibinfo{year}{2021}).
\newblock


\bibitem[Sweeney(2002)]%
        {privacy2}
\bibfield{author}{\bibinfo{person}{Latanya Sweeney}.} \bibinfo{year}{2002}\natexlab{}.
\newblock \showarticletitle{k-anonymity: A model for protecting privacy}.
\newblock \bibinfo{journal}{\emph{International journal of uncertainty, fuzziness and knowledge-based systems}} \bibinfo{volume}{10}, \bibinfo{number}{05} (\bibinfo{year}{2002}), \bibinfo{pages}{557--570}.
\newblock


\bibitem[Touvron et~al\mbox{.}(2021)]%
        {deit}
\bibfield{author}{\bibinfo{person}{Hugo Touvron}, \bibinfo{person}{Matthieu Cord}, \bibinfo{person}{Matthijs Douze}, \bibinfo{person}{Francisco Massa}, \bibinfo{person}{Alexandre Sablayrolles}, {and} \bibinfo{person}{Herv{\'e} J{\'e}gou}.} \bibinfo{year}{2021}\natexlab{}.
\newblock \showarticletitle{Training data-efficient image transformers \& distillation through attention}. In \bibinfo{booktitle}{\emph{International conference on machine learning}}. PMLR, \bibinfo{pages}{10347--10357}.
\newblock


\bibitem[Tramer and Boneh(2021)]%
        {dpmoredata}
\bibfield{author}{\bibinfo{person}{Florian Tramer} {and} \bibinfo{person}{Dan Boneh}.} \bibinfo{year}{2021}\natexlab{}.
\newblock \showarticletitle{Differentially Private Learning Needs Better Features (or Much More Data)}. In \bibinfo{booktitle}{\emph{International Conference on Learning Representations}}.
\newblock
\urldef\tempurl%
\url{https://openreview.net/forum?id=YTWGvpFOQD-}
\showURL{%
\tempurl}


\bibitem[Yousefpour et~al\mbox{.}(2021)]%
        {opacus}
\bibfield{author}{\bibinfo{person}{Ashkan Yousefpour}, \bibinfo{person}{Igor Shilov}, \bibinfo{person}{Alexandre Sablayrolles}, \bibinfo{person}{Davide Testuggine}, \bibinfo{person}{Karthik Prasad}, \bibinfo{person}{Mani Malek}, \bibinfo{person}{John Nguyen}, \bibinfo{person}{Sayan Ghosh}, \bibinfo{person}{Akash Bharadwaj}, \bibinfo{person}{Jessica Zhao}, \bibinfo{person}{Graham Cormode}, {and} \bibinfo{person}{Ilya Mironov}.} \bibinfo{year}{2021}\natexlab{}.
\newblock \showarticletitle{Opacus: {U}ser-Friendly Differential Privacy Library in {PyTorch}}.
\newblock \bibinfo{journal}{\emph{arXiv preprint arXiv:2109.12298}} (\bibinfo{year}{2021}).
\newblock


\bibitem[Yu et~al\mbox{.}(2021)]%
        {dplora1}
\bibfield{author}{\bibinfo{person}{Da Yu}, \bibinfo{person}{Saurabh Naik}, \bibinfo{person}{Arturs Backurs}, \bibinfo{person}{Sivakanth Gopi}, \bibinfo{person}{Huseyin~A Inan}, \bibinfo{person}{Gautam Kamath}, \bibinfo{person}{Janardhan Kulkarni}, \bibinfo{person}{Yin~Tat Lee}, \bibinfo{person}{Andre Manoel}, \bibinfo{person}{Lukas Wutschitz}, {et~al\mbox{.}}} \bibinfo{year}{2021}\natexlab{}.
\newblock \showarticletitle{Differentially private fine-tuning of language models}.
\newblock \bibinfo{journal}{\emph{arXiv preprint arXiv:2110.06500}} (\bibinfo{year}{2021}).
\newblock


\bibitem[Yu et~al\mbox{.}(2020)]%
        {yu2020not}
\bibfield{author}{\bibinfo{person}{Da Yu}, \bibinfo{person}{Huishuai Zhang}, \bibinfo{person}{Wei Chen}, {and} \bibinfo{person}{Tie-Yan Liu}.} \bibinfo{year}{2020}\natexlab{}.
\newblock \showarticletitle{Do not Let Privacy Overbill Utility: Gradient Embedding Perturbation for Private Learning}. In \bibinfo{booktitle}{\emph{International Conference on Learning Representations}}.
\newblock


\bibitem[Zagoruyko and Komodakis(2016)]%
        {wrn}
\bibfield{author}{\bibinfo{person}{Sergey Zagoruyko} {and} \bibinfo{person}{Nikos Komodakis}.} \bibinfo{year}{2016}\natexlab{}.
\newblock \showarticletitle{Wide residual networks}.
\newblock \bibinfo{journal}{\emph{arXiv preprint arXiv:1605.07146}} (\bibinfo{year}{2016}).
\newblock


\end{thebibliography}
\clearpage
\newpage
\appendix
\onecolumn

\section{Experiments}
\subsection{Experimental Setup}
All experiments were carried out on a computing cluster. Experiments were run on an Intel Xeon Gold 6248 machine with 20 CPUs and a single NVIDIA V100 GPU. Our machine is equipped with 170GB of CPU RAM and 32GB of CUDA memory. For our experiments, we use the PyTorch and the Opacus Library \cite{opacus} to implement all neural network models and the DP training.

\subsection{Ablation Study Details}\label{app:ablation}
Here, we discuss the details of the algorithms used in our ablation study in Section~\ref{sec:experiments}. In particular,  we show the workflow for \averagednoisy~in Algorithm~\ref{algo:ex1} and the workflow for \averagedtrue~in Algorithm~\ref{algo:ex2}. Generally speaking, these two algorithms follow the same workflow as Algorithm~\ref{algo:row-pruning}, but the mask selection is different for each of them. \averagedtrue~uses the true gradients, which implies the algorithm will not satisfy DP, but this allows us to understand how good our mask selection procedure is. \averagednoisy~uses DP-SGD gradients for mask selection, which allows us to understand how important grouping is in \ourmethod.

\begin{algorithm}[!h]
    \caption{Selection via \averagednoisy}
    \label{algo:ex1}
    \begin{algorithmic}
        \State \textbf{Input:} $\B W_\text{old}$ (pre-trained weights), $T_0=10$ (mask finding epoch), $T=50$(\#epochs)
        \State Train $\B W_{\text{old}}$ for $T_0$ epochs with DP-SGD.
        \State $\tilde{\B u}=\B{0}$
        \For{each Poisson-sampled batch $B_t$ ($1$ epoch)}
                \State $\tilde{\B u} \pluseq \sum_{i\in B_t} \B g^{t,i}/\max\left\{1,\|\B g^{t,i}\|_2/C\right\} + \mathcal{N}(0, C^2\sigma^2 \B I) $ 
               \State \Comment{Absolute value of DP-SGD gradients}
        \EndFor
        \State $\hat{\B m}=\argtopk(\tilde{\B u})$ \Comment{Select the sparse mask}
        \State Continue training $\B{W}\mapsto \mathcal{L}(\B{W}_{\text{old}}+\hat{\B m}\odot \B{W} )$ with DP-SGD for $T - T_0 - 1$ epochs.
        \State \Comment{Sparse Fine-Tuning, use Clipping $C$ and noise variance $\sigma^2$.}
    \end{algorithmic}
\end{algorithm}

\begin{algorithm}[!h]
    \caption{\averagedtrue}
    \label{algo:ex2}
    \begin{algorithmic}[]
        \State \textbf{Input:} $\B W_\text{old}$ (pre-trained weights), $T_0=10$ (mask finding epoch), $T=50$(\#epochs)
        \State Train $\B W_{\text{old}}$ for $T_0$ epochs with DP-SGD.
        \State $\tilde{\B u}=\B{0}$
        \For{each Poisson-sampled batch $B_t$ ($1$ epoch)}
                \State $\tilde{\B u} \pluseq \sum_{i\in B_t} \abs{\B g^{t,i}}$
               \State \Comment{Averaged true gradients}
        \EndFor
        \State $\hat{\B m}=\argtopk(\tilde{\B u})$ \Comment{Select the sparse mask (non-privatized selection)}
        \State Continue training $\B{W}\mapsto \mathcal{L}(\B{W}_{\text{old}}+\hat{\B m}\odot \B{W} )$ with DP-SGD for $T - T_0 - 1$ epochs.
        \State \Comment{Sparse Fine-Tuning, use Clipping $C$ and noise variance $\sigma^2$.}
    \end{algorithmic}
\end{algorithm}

\begin{algorithm}[!h]
    \caption{\mpweights}
    \label{algo:mpweights-alg}
    \begin{algorithmic}[]
        \State \textbf{Input:} $\B W_\text{old}$ (pre-trained weights), $T=50$(\#epochs)
        \State $\hat{\B m}=\argtopk(|\tilde{\B  W_\text{old}}|)$
        \State \Comment{Select the sparse mask using the magnitude of public pre-trained weights $ \B W_\text{old}$}
        \State Train $\B{W}\mapsto \mathcal{L}(\B{W}_{\text{old}}+\hat{\B m}\odot \B{W} )$ with DP-SGD for $T$ epochs.
        \State \Comment{Sparse Fine-Tuning, use Clipping $C$ and noise variance $\sigma^2$.}
    \end{algorithmic}
\end{algorithm}

\subsection{Effect of $\varepsilon$ on Accuracy/Percentage of trainable parameters profile}
One motivation of sparse fine-tuning in the context of DP-SGD is that each trainable parameter is updated with an independent (to other parameters) noise. In this subsection, we further explore this idea by demonstrating the profile of Accuracy/Percentage of trainable parameters for the DeiT Tiny network for $\varepsilon = 2$ and $\varepsilon = 8$. It shows that in a lower signal-to-noise ratio, with $\varepsilon = 2$ corresponding to a high privacy regime, it seems more suitable to use a smaller number of trainable parameters compared to a high signal-to-noise ratio, with $\varepsilon = 8$ corresponding to a moderate privacy regime. Our explorations have shown that pushing these boundaries a lot further would lead to DP-BitFit and All layer fine-tuning to out-perform mask selection approaches (including \averagedtrue which has an infinite privacy budget for mask selection) for extremely low and extremely high (approaching non-private regime) $\varepsilon$ values respectively.

\begin{figure}
    \includegraphics[width=0.48\textwidth]{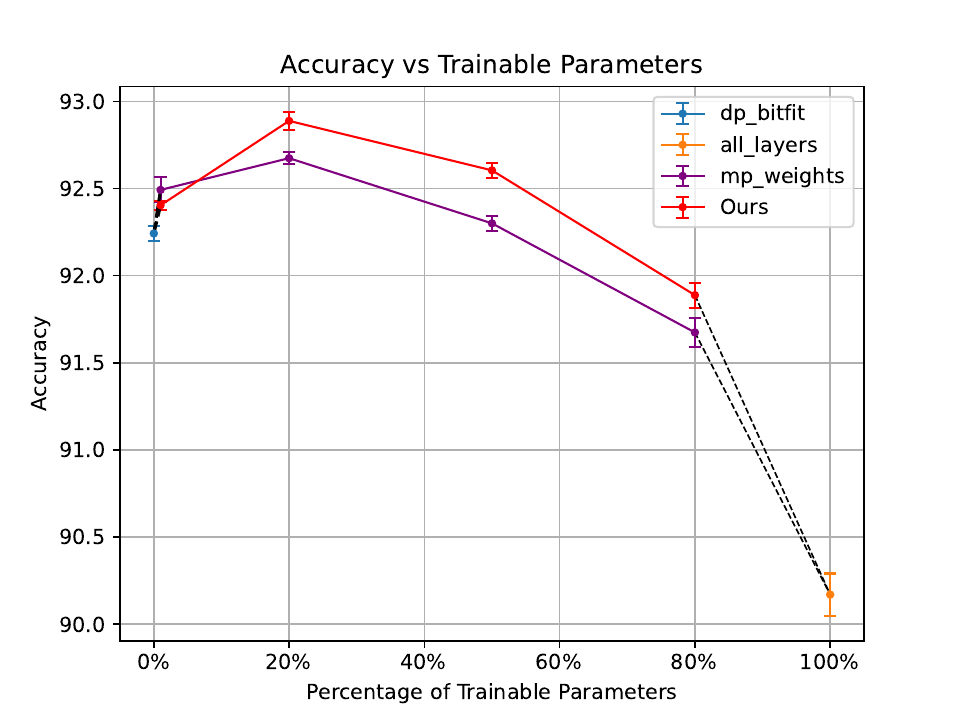}
    \includegraphics[width=0.48\textwidth]{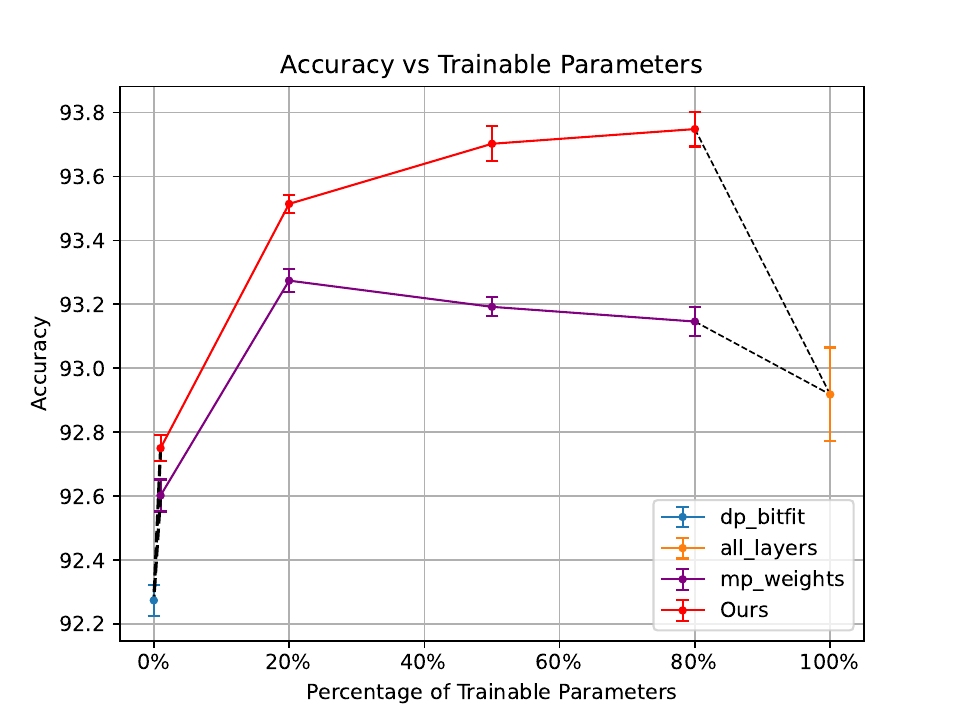}
    \caption{Profile of Accuracy/Percentage of trainable parameters for DeiT Tiny under $(\varepsilon, \delta) = (2, 10^{-5})$ (left) and $(\varepsilon, \delta) = (8, 10^{-5})$ (right) DP-guarantees.}
    \label{fig:tiny_interpolation}
\end{figure}

\end{document}